\documentclass{article}

\usepackage{microtype}
\usepackage{graphicx}
\usepackage{booktabs} 
\usepackage[T1]{fontenc}

\usepackage{hyperref}

\usepackage{url}            
\usepackage{booktabs}       
\usepackage{amsfonts}       
\usepackage{nicefrac}       
\usepackage{microtype}      
\usepackage{lipsum}		
\usepackage{natbib}
\usepackage{doi}
\usepackage{amsmath, amssymb}
\usepackage{soul, xcolor}

\usepackage{caption}
\usepackage{subcaption}

\usepackage[accepted]{icml2023}

\usepackage{amsmath}
\usepackage{amssymb}
\usepackage{mathtools}
\usepackage{amsthm}

\usepackage[capitalize,noabbrev]{cleveref}

\theoremstyle{plain}
\newtheorem{theorem}{Theorem}[section]

\theoremstyle{definition}

\theoremstyle{remark}
\newtheorem{remark}[theorem]{Remark}

\usepackage[textsize=tiny]{todonotes}

\icmltitlerunning{Trajectory Generation, Control, and Safety with Denoising Diffusion Probabilistic Models}

\begin{document}

\twocolumn[
\icmltitle{Trajectory Generation, Control, and Safety with Denoising Diffusion Probabilistic Models}




\begin{icmlauthorlist}
\icmlauthor{Nicolò Botteghi}{yyy}
\icmlauthor{Federico Califano}{xxx}
\icmlauthor{Mannes Poel}{zzz}
\icmlauthor{Christoph Brune}{yyy}
\end{icmlauthorlist}

\icmlaffiliation{yyy}{Mathematics of Imaging and AI, University of Twente, Enschede, The Netherlands}
\icmlaffiliation{xxx}{Robotics and Mechatronics, University of Twente, Enschede, The Netherlands}
\icmlaffiliation{zzz}{Datamanagement and Biometrics, University of Twente, Enschede, The Netherlands}

\icmlcorrespondingauthor{Nicolò Botteghi}{n.botteghi@utwente.nl}

\icmlkeywords{Denoising Diffusion Probabilistic Models, Control Barrier Functions, Offline Model-based Reinforcement Learning, Robotics, Safety-critical Planning}

\vskip 0.3in
]



\printAffiliationsAndNotice{}  

\begin{abstract}
We present a framework for safety-critical optimal control of physical systems based on denoising diffusion probabilistic models (DDPMs). The technology of control barrier functions (CBFs), encoding desired safety constraints, is used in combination with DDPMs to plan actions by iteratively denoising trajectories through a CBF-based guided sampling procedure. At the same time, the generated trajectories are also guided to maximize a future cumulative reward representing a specific task to be optimally executed.
The proposed scheme can be seen as an offline and model-based reinforcement learning algorithm resembling in its functionalities a model-predictive control optimization scheme with receding horizon in which the selected actions lead to optimal and safe trajectories. 
\end{abstract}

\section{Introduction}\label{sec:intro}

When controlling a physical system, the concept of safety is crucial. In general, safety is not a system theoretic property (like e.g., stability) which possesses a precise mathematical definition, and safety requirements depend on the specific safety hazards that are considered. Systems for which some instance of safety is considered of paramount importance, are referred to as safety-critical systems. Examples include autonomous driving with adaptive cruise control and human-collision avoidance for collaborative robots \cite{Ames2019ControlApplications}. 
Safety constraints are conceptually distinct from the task-based specifications. 
For example, a collaborative robot performing a trajectory tracking task might be controlled using Lyapunov-based designs or learned control policies, but the underlying safety requirement of avoiding collisions with humans is transversal to any chosen controller, and must be always satisfied. 

Control barrier functions (CBFs) \cite{Ames2017ControlSystems,Ames2019ControlApplications} represent a formal framework aiming to achieve safety as a hard constraint in an optimization problem in which the cost function encodes information on the nominal task to be executed. In particular CBF-based safety constraints are represented by forward invariance of so-called safe sets, i.e. subsets of the state space which the controlled system should not leave during the task execution. We stress that within this context, safety becomes a mathematically rigorous system theoretic property and, even if unable to represent any possible safety hazard, it is very useful to design safety constraints, e.g. kinematic constraints for mechanical systems \cite{Singletary2021Safety-CriticalSystems}.    

Reinforcement learning (RL) \cite{Sutton2018ReinforcementIntroduction} has achieved outstanding success in control of dynamical and physical systems directly from high-dimensional sensory data \cite{Kaelbling1996ReinforcementSurvey, Kaiser2019Model-BasedAtari, Li2017DeepOverview, Botteghi2022UnsupervisedReview}. RL algorithms can be classified as either model-free or model-based. Model-free RL algorithms learn optimal policies directly from observations and rewards received by the agent from the environment after each action taken. On the other side, model-based algorithms first learn the environment dynamics, i.e. forward and reward models, and then use the learned models to generate samples for optimizing the policy. Independently of the chosen method, RL lacks of safety guarantees, which is one of the reasons why learning schemes are still poorly used in real safety-critical applications. As a consequence, embedding safety guarantees, for example using CBFs, in learning schemes is becoming a major research topic in the learning community, see \cite{Dawson2023SafeControl,Cheng2019End-to-EndTasks,Ma2022JointLearning} and references therein.

\begin{figure*}[ht!]
    \centering
    \includegraphics[page=1, width=\textwidth]{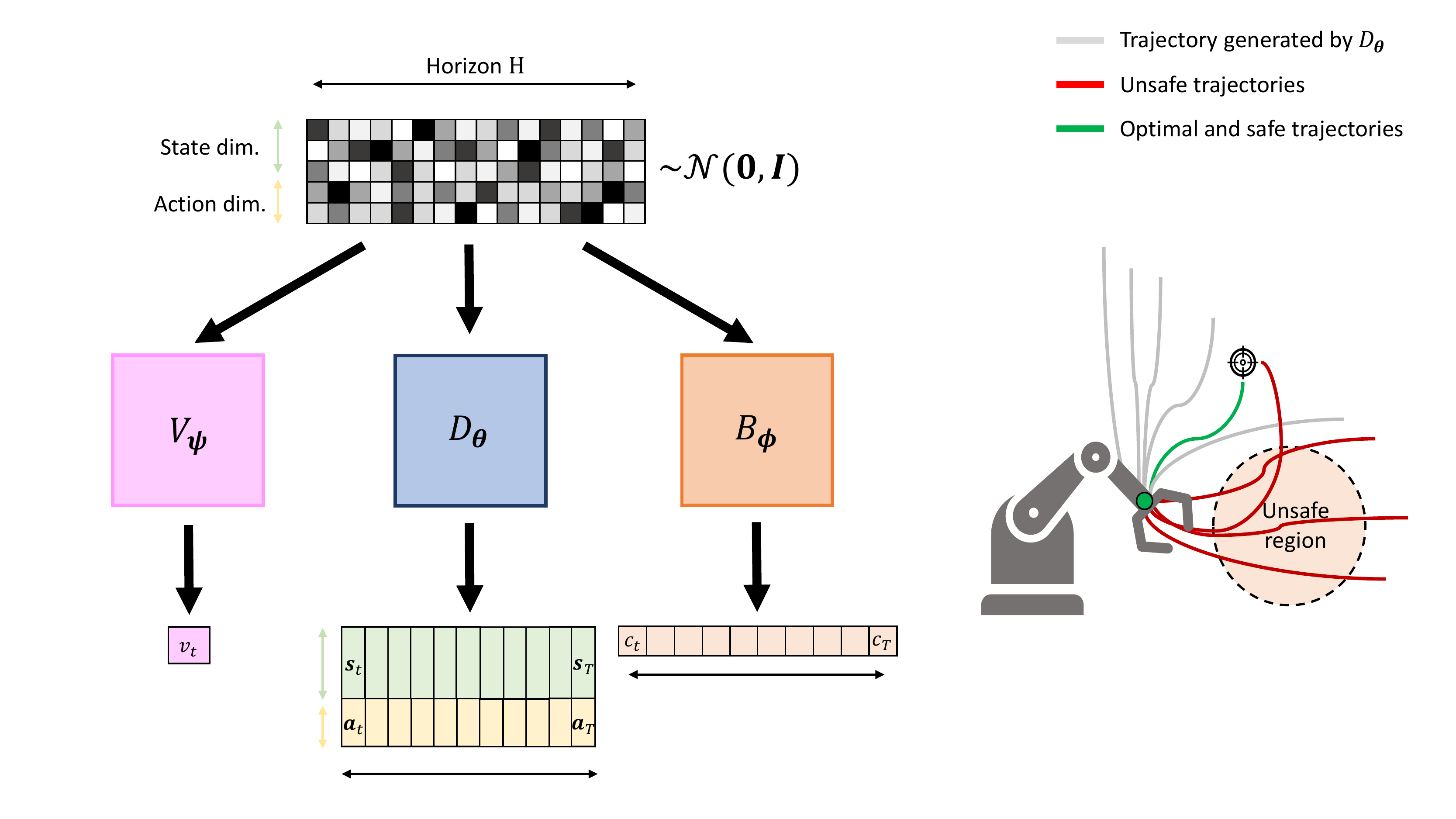}
    \caption{Proposed framework for planning and control using denoising diffusion probabilistic models. Generation from random noise of dynamically consistent state-action trajectories with horizon H from a generic timestep $t$ to $T = t+H$.}
    \label{fig:framework}
\end{figure*}

Intrinsically related to safety and CBFs, we find the need for a forward dynamical model able to accurately predict the evolution over time of the system given current state and control inputs. The forward model can be either derived from physical governing laws, especially for simple dynamical systems, or learned from data when limited or no physical knowledge is available \cite{Brunton2022Data-DrivenControl}. The latter is often the case for complex dynamical systems \cite{Brunton2022Data-DrivenControl} and the progress of deep learning \cite{Lecun2015DeepLearning} has made learning, and controlling, dynamical systems from data increasingly popular in recent years. In most cases, data-driven methods focus on learning, for example via neural networks or Gaussian processes \cite{Rasmussen2004GaussianLearning}, a (Markovian) 1-step-ahead forward model predicting the state at the next timestep given (history of) state(s) and action(s). However, when making long-term prediction, the 1-step-ahead forward model has to be iteratively called at the price of accumulating more and more prediction error the longer the horizon. An alternative solution was proposed by \cite{Janner2022PlanningSynthesis}, where instead of learning a model predicting the next state, they introduce a trajectory-based generative model outputting sequences of state-action pairs, i.e. trajectories. It is worth mentioning that this model is only locally Markovian since each step of the trajectory is a function of the past and future state-action pairs. Therefore, the model has to be able to learn the causal relations from data, i.e. observed state-action pairs over time, to properly generate trajectories. 


With reference to Figure \ref{fig:framework}, we frame our contribution in the context of safety-critical optimal control in an offline and model-based RL framework, proposing a scheme utilizing denoising diffusion probabilistic models (DDPMs) \cite{Sohl-Dickstein2015DeepThermodynamics, Ho2020DenoisingModels} for optimal and safe trajectory generation. As hallmark of our contribution, three properties are approximated by three different DDPMs, and as such represent conceptually distinct units which can be independently changed in the proposed architecture:
\begin{itemize}
    \item \textit{dynamically consistent trajectory generation}, i.e. a diffusion model $D_{\boldsymbol{\theta}}$ is trained to generate trajectories consistent with the dynamics of the system,
    \item \textit{value estimation}, i.e. a diffusion model $V_{\boldsymbol{\psi}}$ is trained to learn the expected return of each trajectory, encoding the specific task to be optimally executed, and
    \item \textit{safety classification}, i.e., a diffusion model $B_{\boldsymbol{\phi}}$ is trained to classify the generated trajectory as safe or unsafe, based on the CBF constraint.
\end{itemize}
Given these three DDPMs, the problem of safety-critical optimal control can be rephrased as a problem of conditional sampling of trajectories, i.e. sequence of state-action pairs, from $D_{\boldsymbol{\theta}}$ using $V_{\boldsymbol{\psi}}$ and $B_{\boldsymbol{\phi}}$ as guides. With this method, the generated trajectories are not only optimal, but also safe. 

Our approach closely relates to the work in \cite{Janner2022PlanningSynthesis}, where two diffusion models, predicting trajectories and values, were proposed together with a conditional sampling scheme. However, differently from them we focus on safety-critical control with the addition of a third diffusion model and a new conditional sampling scheme combining optimality and safety.
After applying the first action of the trajectory, a new conditional sampling step takes place given the next state of the environment as initial state, resembling an optimization scheme with receding horizon. As a consequence, similarly to what has been achieved in \cite{Zeng2021Safety-CriticalFunction} in a model predictive control (MPC) framework, the CBF-based safety constraint influences the choice of the trajectory over the whole planning horizon, and not only at the current state, making the planning appoach not over-conservative. The view on the whole trajectory is the main conceptual differences with respect to \cite{Cheng2019End-to-EndTasks}, also combining RL and CBFs, but where the CBF-based constraint is solved at every iteration to modify online the model-free RL policy. 



\section{Preliminaries}\label{sec:preliminaries}

\subsection{Control Barrier Functions}\label{subsec:control_barrier_functions}
Control barrier functions (CBFs) are used to guarantee invariance of a set $\mathcal{C}$, normally referred to as \textit{safe set}. The latter is designed to encode constraints depending on the state of the controlled system, e.g., obstacle avoidance for robot manipulators. We present the relevant background directly in its discrete-time formulation \cite{Ahmadi2019SafeFunctions,Zeng2021Safety-CriticalFunction,Xiong2023Discrete-TimeCase} to conveniently integrate the technique in the developed learning scheme. Furthermore, we use the standard RL notation to indicate state, action, rewards, and value function \cite{Sutton2018ReinforcementIntroduction}.

Consider a discrete-time system 

\begin{equation}
\label{eq:system}
\mathbf{s}_{t+1}=f(\mathbf{s}_t,\mathbf{a}_t) 
\end{equation}
where $\mathbf{s}_t\in \mathcal{S}\subset  \mathbb{R}^n$ is the state at time step $t\in \mathbb{N}$, the control action $\mathbf{a}_t\in \mathcal{A}$ is applied to the system, and $f: \mathcal{S} \times \mathcal{A} \to \mathcal{S}$ is a continuous map. The safe set $\mathcal{C}$ is defined as
\begin{align*}
    \mathcal{C} = \{ \mathbf{s}\in \mathcal{S} : h(\mathbf{s})\geq0 \}, \\
     \partial \mathcal{C} = \{ \mathbf{s}\in \mathcal{S} : h(\mathbf{s})=0 \}, \\
      \textrm{Int} (\mathcal{C}) = \{ \mathbf{s}\in \mathcal{S} : h(\mathbf{s})>0 \},
\end{align*}
where $h:\mathcal{S}\to \mathbb{R}$ is a continuous map. Remind that a class $\mathcal{K}$ function is a strictly increasing function $\alpha:[0,a) \to [0,\infty)$ with $\alpha(0)=0$ and $a>0$. The function $h(\mathbf{s})$ is then defined to be a CBF if there exists a class $\mathcal{K}$ function $\alpha$ satisfying $\alpha(u)<u, \forall u>0$ such that

\begin{equation}
\label{eq:CBFcondition}
    h(\mathbf{s}_{t+1})-h(\mathbf{s}_{t})\geq -\alpha(h(\mathbf{s}_t)), \,\,\,\, \forall \mathbf{s} \in \mathcal{S}.
\end{equation}

The set $\mathcal{C}$ is called forward invariant if $\mathbf{s}_0 \in \mathcal{C}$ implies $\mathbf{s}_t \in \mathcal{C}, \forall t \in \mathbb{N}$. The following theorem, which represents the core of  CBFs in the current context, relates existence of CBF and forward invariance of $\mathcal{C}$ achieved by means of control.

\begin{theorem}\cite{Ahmadi2019SafeFunctions}
\label{th:CBF}
Consider system (\ref{eq:system}) and the previously defined set $\mathcal{C}$. Any control policy $\pi(\mathbf{a}_t|\mathbf{s}_t)$ implementing $\mathbf{a}_t$ in a way that (\ref{eq:CBFcondition}) is satisfied will render $\mathcal{C}$ forward invariant.
\end{theorem}

\begin{remark}
We stress that in the literature different types of CBFs are present. These vary, both in continuous and discrete time, on the basis of i) the specific set $\mathcal{S}$ where condition (\ref{eq:CBFcondition}) must hold, which can be a proper subset of the total state space (in this case the CBFs are normally referred to explicitly as "CBFs on $\mathcal{S}$"); and ii) whether the CBF diverges or goes to $0$ at the boundary $\partial \mathcal{C}$ of the safe set. In the latter case, which is the one treated in this work, CBFs are sometimes referred to as "zeroing" CBFs, and have the advantage to be well defined outside the safe set. It follows (see proof Theorem 1 in \cite{Ahmadi2019SafeFunctions}) that existence of zeroing CBFs on $\mathcal{S}$ not only achieve forward invariance, but also asymptotic stability of $\mathcal{C}$.
\end{remark}

\begin{remark}
Condition (\ref{eq:CBFcondition}) for an autonomous discrete time system $\mathbf{s}_{t+1}=f(\mathbf{s}_t)$ is actually necessary and sufficient (i.e., equivalent) to achieve forward invariance of the set $\mathcal{C}$ (see Theorem 1 in \cite{Ahmadi2019SafeFunctions}). It follows that finding a control policy applied to (\ref{eq:system}) which satisfies (\ref{eq:CBFcondition}) is also necessary and sufficient (i.e., equivalent) to produce forward invariance of $\mathcal{C}$ in the closed-loop system. An interesting consequence is that the discrete time CBF approach is not conservative (but in fact equivalent) to achieve forward invariance of a set.
\end{remark}

\begin{remark}
Normally one chooses the linear class $\mathcal{K}$ function $\alpha(u)=\lambda u$, with $\lambda\in \mathbb{R}, \,\,0 \leq \lambda \leq1$. In this way it holds $h(\mathbf{s}_{t})\geq(1-\lambda)^t h(\mathbf{s}_{0})$, which clearly shows forward invariance of $\mathcal{C}$ for every admissible $\lambda$. The constraint (\ref{eq:CBFcondition}) with the choice $\lambda=1$ corresponds to achive invariance in the less conservative case, i.e., $h(\mathbf{s}_t)\geq 0, \forall t\in \mathbb{N}$, while the choice $\lambda=0$ collapses (\ref{eq:CBFcondition}) into a Lyapunov condition achieving $h(\mathbf{s}_t)\geq h(\mathbf{s}_{t-1}), \forall t \in \mathbb{N}$.
\end{remark}

The way discrete-time CBFs are implemented in practice is by casting condition (\ref{eq:CBFcondition}) as a constraint of a discrete optimisation problem, which in its basic forms can be represented as

\begin{equation}
\label{eq:LQ}
\begin{aligned}
\mathbf{a}^*_t&=\textrm{argmin}_{\mathbf{a}_t\in \mathcal{A}} \quad  \mathbf{J}(\mathbf{s}_t,\mathbf{a}_t)\\
&\textrm{s.t.}\quad h(f(\mathbf{s}_t,\mathbf{a}_t))-h(\mathbf{s}_{t})\geq -\alpha(h(\mathbf{s}_t))
\end{aligned}
\end{equation}

The level of sophistication of the latter can vary on the basis of the application, by adding other constraints (e.g., input constraints), and/or slack variables to help the feasibility of the problem.

\subsection{Denoising Diffusion Probabilistic Models}\label{subsec:DDPM}

Denoising diffusion probabilistic models (DDPMs) \cite{Sohl-Dickstein2015DeepThermodynamics, Ho2020DenoisingModels} are a class of (deep) probabilistic generative models, such as generative adversarial networks \cite{Goodfellow2020GenerativeNetworks}, variational autoencoders \cite{Kingma2014Auto-encodingBayes}, and score-matching models \cite{Song2019GenerativeDistribution, Song2020ImprovedModels} learning generative data distributions out of which we can sample new data. DDPMs are parametrized Markov chains generating data from random noise by performing a step-wise denoising of the random vectors. The denoising process, often referred to as the reverse process, is learned from data with the goal of reverting the forward process of gradually adding noise to the data. The forward process can be written as:
\begin{equation}
\begin{split}
q(\mathbf{x}^{1:K}|\mathbf{x}^0) &= \prod_{k=1}^Kq(\mathbf{x}^k|\mathbf{x}^{k-1}), \\
q(\mathbf{x}^k|\mathbf{x}^{k-1})&=\mathcal{N}(\sqrt{1-\beta_k}\mathbf{x}^{k-1}, \beta_k\mathbf{I})
\end{split}
    \label{eq:forward_process}
\end{equation}
where $\mathbf{x} \sim \mathcal{D}$ is a data sample from a dataset $\mathcal{D}$, the superscript $k=1, \dots, K$ indicates the $k^{\text{th}}$ diffusion step, $K$ the number of diffusion steps, $\mathbf{x}^0$ the noise-free data, and $\beta_k \in (0, 1)$ is an hyperparameter controlling the variance of the forward diffusion process. The reverse process is described by:
\begin{equation}
\begin{split}
    p_{\boldsymbol{\theta}}(\mathbf{x}^{0:K}) &= \prod_{k=1}^Kp_{\boldsymbol{\theta}}(\mathbf{x}^{k-1}|\mathbf{x}^k) \\
    p_{\boldsymbol{\theta}}(\mathbf{x}^{k-1}|\mathbf{x}^k)&=\mathcal{N}(\boldsymbol{\mu}_{\boldsymbol\theta}(\mathbf{x}^k, k), \boldsymbol{\Sigma}^k)
\end{split}
    \label{eq:reverse_process}
\end{equation}
where $\boldsymbol{\mu}_{\boldsymbol\theta}(\mathbf{x}^k, k)$ is the mean learned by the diffusion model, $\boldsymbol{\Sigma}^k$ is the covariance matrix usually following a cosine schedule \cite{Nichol2021ImprovedModels} (and not learned from data), and the subscript $\boldsymbol{\theta}$ indicates the (learnable) parameters of the diffusion model.

DDPMs can be trained by maximizing the variational lower bound similarly to variational autoencoders \cite{Sohl-Dickstein2015DeepThermodynamics}, or using the simplified objective introduced by \cite{Ho2020DenoisingModels}:
\begin{equation}
\begin{split}
L(\boldsymbol{\theta}) &= \mathbb{E}_{k, \epsilon, \mathbf{x}_0} [||\epsilon^k - \epsilon_{\boldsymbol{\theta}}(\mathbf{x}^k, k) ||^2] \\
&= \mathbb{E}_{k, \epsilon, \mathbf{x}_0} [||\epsilon^k - \epsilon_{\boldsymbol{\theta}}(\sqrt{\Bar{\alpha_k}}\mathbf{x}^0+\sqrt{1-\Bar{\alpha}_k}\epsilon^k, k) ||^2]
\end{split}
    \label{eq:diffusion_simple_objective}
\end{equation}
where $k \sim \mathcal{U}\{1, \dots, K \}$ with $\mathcal{U}$ indicating a uniform distribution, $\epsilon^k \sim \mathcal{N}(\mathbf{0}, \mathbf{I})$ is the target noise, $\Bar{\alpha}_k=\prod_{k=1}^K\alpha_k$ with $\alpha_k=1-\beta_k$, and with the learnable mean rewritten as:
\begin{equation}
\boldsymbol{\mu}_{\boldsymbol{\theta}}(\mathbf{x}^k, k)=\frac{1}{\sqrt{\alpha}_k}(\mathbf{x}^k-\frac{1-\alpha_k}{\sqrt{1-\Bar{\alpha}_k}}\epsilon_{\boldsymbol{\theta}}(\mathbf{x}^k, k))
    \label{eq:mu_rewritten}
\end{equation}

To further improve and control the generation process, it is possible to introduce a classifier $p_{\boldsymbol{\phi}}(y|\mathbf{x}^k)$ and use its gradient $\nabla_{\mathbf{x}^k}p_{\boldsymbol{\phi}}(y|\mathbf{x}^k)$  with respect to the input $\mathbf{x}^k$ to guide the reverse diffusion process towards samples from a class $y$. This procedure is commonly referred to as classifier-guided sampling \cite{Dhariwal2021DiffusionSynthesis}.

\subsubsection{Planning and Control}

Instead of considering a generic data $\mathbf{x}$, we assume we have recorded trajectories of our dynamical system $\boldsymbol{\tau}_t=(\mathbf{s}_t, \mathbf{a}_t, \dots, \mathbf{s}_T, \mathbf{a}_T)$, where $\mathbf{s}_t$ corresponds to the state vector at timestep $t$, $\mathbf{a}_t$ to the control action, and the subscript $t\in \{0,...,T \}$ to the actual timestep of agent-environment interaction of the underlying RL scheme, while the superscript $k$ indicates the diffusion steps as introduced in Section \ref{subsec:DDPM}. Given a sampled trajectory $\boldsymbol{\tau}^K\sim \mathcal{N}(\mathbf{0}, \mathbf{I})$, it is possible to use a trained diffusion model to reverse the forward process and to generate a trajectory $\boldsymbol{\tau}_t$ consistent with the observed dynamics of the system, as shown in \cite{Janner2022PlanningSynthesis}:
\begin{equation}
p_{\boldsymbol{\theta}} (\boldsymbol{\tau}_t^{k-1}|\boldsymbol{\tau}_t^{k}) = \mathcal{N}(\boldsymbol{\mu}_{\boldsymbol\theta}(\boldsymbol{\tau}_t^{k}, k), \boldsymbol{\Sigma}^k)
    \label{eq:reverse_process_trajectory}
\end{equation}
In a classic model-based RL fashion, one may sample trajectories from the reverse diffusion model, apply the control actions to the real environment and evaluate the value of each visited state. However, analogously to the classifier-guided sampling introduced by \cite{Dhariwal2021DiffusionSynthesis}, diffusion models can be used to perform conditional sampling, e.g., sampling images from a specific class, by using a classifier to adjust the mean of the reverse process during the diffusion steps. The use of conditional sampling for control was first explored by \cite{Janner2022PlanningSynthesis} by transforming the problem of optimal trajectory generation in a problem of conditional sampling:
\begin{equation}
    \Tilde{p}_{\boldsymbol{\theta}}(\boldsymbol{\tau}) \approx p_{\boldsymbol{\theta}}(\boldsymbol{\tau})f(\boldsymbol{\tau})
\end{equation}
where $f(\boldsymbol{\tau}_t)$ may encode prior information, desired goals, or general reward and cost functions to optimize. In particular, we can use a trained value function model, indicated by $ V_{\boldsymbol{\psi}}$ to sample trajectories maximizing the value function, i.e. optimal trajectories. A single guided-diffusion step can be written as:
\begin{equation}
    \boldsymbol{\tau}^{k-1}\sim \mathcal{N}( \boldsymbol{\mu}^{k-1}_{\boldsymbol{\theta}}+\eta_1\boldsymbol{\Sigma}^{k-1}\nabla_{\boldsymbol{\tau}^k} V_{\boldsymbol{\psi}}(\boldsymbol{\mu}_{\boldsymbol{\theta}}), \boldsymbol{\Sigma}^{k-1})
    \label{eq:planning_diffusion}
\end{equation}
with $\eta_1$ a scaling constant, and $\boldsymbol{\mu}_{\boldsymbol{\theta}}=\boldsymbol{\mu}_{\boldsymbol\theta}(\boldsymbol{\tau}_t^{k}, k)$ for the sake of lightening the notation.
After the generation the planning, the first action is executed in the environment, and similarly to MPC methods with receding horizon, the plan is generated again given the next state of the environment as initial state of the trajectory.

While this planning strategies allows the generation of optimal trajectories, differently from \cite{Janner2022PlanningSynthesis}, in our work we focus on the problem of optimal planning under safety constraint where we want to generate not only optimal, but also safe trajectories. 

\section{Optimal and Safe Trajectory Generation with Denoising Diffusion Probabilistic Models}\label{sec:methodology}

In this work, we study the problem optimal and safe trajectory generation using DDPMs and conditional sampling. With reference to Figure \ref{fig:framework}, we decompose the problem of optimal and safe trajectory planning into the problem of learning:
\begin{itemize}
    \item $D_{\boldsymbol{\theta}}$ generating trajectories $\boldsymbol{\tau}_t=(\mathbf{s}_t, \mathbf{a}_t, \dots, \mathbf{s}_T, \mathbf{a}_T)$ from random noise consistent with the measurement data of the dynamical system considered, 
    \item $V_{\boldsymbol{\psi}}$ estimating the value, i.e. expected discounted return, of each trajectory, again from random noise, and
    \item $B_{\boldsymbol{\phi}}$ classifying whether each timestep of the trajectories generated is safe or unsafe in accordance with the CBF-based safety constraint, defined by Equation \eqref{eq:CBFcondition}.
\end{itemize}
These three models are DDPMs (see Section \ref{subsec:DDPM}) with architecture described in details in Appendix \ref{app:Architecture}. 

While it is possible to train a single DDPM to generate trajectories, predicting values, and assessing safety of the state-action pairs, the independence of the three models is important for data efficiency and flexibility of use. For example, if a robot is moved to a new room the safety regions may change, while its dynamics or task may not. With our framework, we just need to retrain the safety classifier $B_{\boldsymbol{\phi}}$. The same holds if we change reward function, with the difference that we need now to retrain only the value function model $V_{\boldsymbol{\psi}}$.

\subsection{Planning as Conditional Sampling}

\begin{figure}[h!]
    \centering
    \includegraphics[page=2, width=\linewidth]{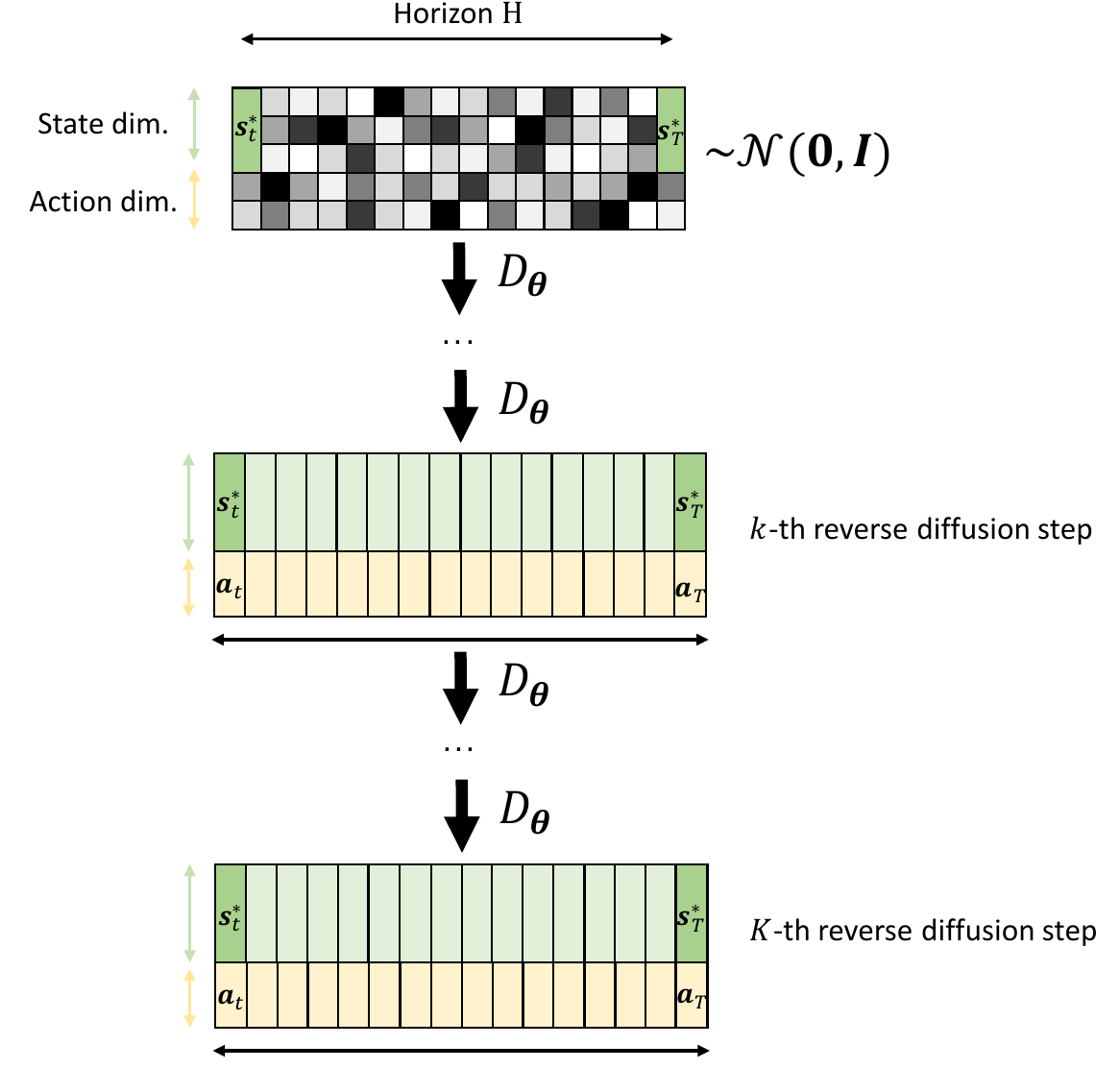}
    \caption{Reverse diffusion process with inpainting of initial and final states of the trajectory indicated with $\mathbf{s}_t^*$ and $\mathbf{s}_T^*$.}
    \label{fig:inpainting}
\end{figure}

In this section, we describe the conditional sampling procedure for the generation of optimal and safe trajectories. Given $D_{\boldsymbol{\theta}}$, we can generate trajectories from random noise accordingly to Equation \eqref{eq:reverse_process_trajectory}. The generation of the trajectories can be conditioned on initial and/or final states of the trajectories through the process called inpainting \cite{Sohl-Dickstein2015DeepThermodynamics}, i.e. the problem of consistently filling the gaps or a trajectory given some known states or actions (see Figure \ref{fig:inpainting} for an example).
Since we focus on path planning with unknown target location, in our experiments we condition the generated plan only on the initial state. Using $V_{\boldsymbol{\psi}}$, we can guide $D_{\boldsymbol{\theta}}$ toward the generation of high-value trajectories according to Equation \eqref{eq:planning_diffusion}, i.e. trajectories reaching the target \cite{Janner2022PlanningSynthesis}, yet possibly unsafe. The guided planning procedure is described in Algorithm \ref{alg:optsafeplan}, where the guide function is generically indicated by $f$.
\begin{algorithm}[tb]
   \caption{Guided Planning}
   \label{alg:optsafeplan}
\begin{algorithmic}
   \STATE {\bfseries Input:} $D_{\boldsymbol{\theta}}$, and guide $f$
    \FOR{$i=1$ {\bfseries to} $I$}
   \STATE Reset environment and sample $\mathbf{s}_1$
   \FOR{$j=1$ {\bfseries to} $J$}
   \STATE Sample plan $\boldsymbol{\tau}^K\sim \mathcal{N}(\mathbf{0}, \mathbf{I})$ conditioned on $\mathbf{s}_j$
   \FOR{$k=K$ {\bfseries to} $1$}
   \STATE $\boldsymbol{\mu}_{\boldsymbol{\theta}}\leftarrow D_{\boldsymbol{\theta}}(\boldsymbol{\tau}^k)$ 
   \STATE $\boldsymbol{\tau}^{k-1}\leftarrow \sim \mathcal{N}(\boldsymbol{\mu}_{\boldsymbol{\theta}}+\eta\boldsymbol{\Sigma}\nabla f, \boldsymbol{\Sigma}^k)$
   \STATE $\boldsymbol{\tau}_{\mathbf{s}_j}^{k-1}\leftarrow \mathbf{s}_j$
   \ENDFOR
   \STATE Apply first action of the plan to the environment
    \STATE  Observe next state $\mathbf{s}_{j+1}$, reward $r_j$, terminal condition $d_{j+1}$, and safety constraint $c_j$
    \STATE $\mathbf{s}_{j}\leftarrow \mathbf{s}_{j+1}$
   \ENDFOR
   \ENDFOR
\end{algorithmic}
\end{algorithm}

To make these high-value trajectories safe, accordingly to the CBFs methodology, one may solve the constrained optimization problem in Equation \eqref{eq:LQ} for each step of the trajectory in order to "filter" the actions. However, the problem in \eqref{eq:LQ} is expensive to solve, especially for non-affine dynamics and/or constraints \cite{Cheng2019End-to-EndTasks}, and has to be computed every time before applying an action to the environment. 

Here, we propose an alternative solution. Recall that $B_{\boldsymbol{\phi}}$ is able to classify whether each step of a trajectory is safe or not. Similarly to the classifier-guide sampling method \cite{Dhariwal2021DiffusionSynthesis}, often employed in the process of generating high-quality images, we can guide $D_{\boldsymbol{\theta}}$ to generate only trajectories belonging to the "safe" class by adjusting the mean of the reverse process accordingly to:
\begin{equation}
    \boldsymbol{\tau}^{k-1} \sim \mathcal{N}( \boldsymbol{\mu}^{k-1}_{\boldsymbol{\theta}}+\eta_2\boldsymbol{\Sigma}^{k-1}\nabla_{\boldsymbol{\tau}^k} \log p_{\boldsymbol{\phi}}(\mathbf{c}|\boldsymbol{\tau}^k), \boldsymbol{\Sigma}^{k-1})
    \label{eq:safe_planning_diffusion}
\end{equation}
where $\eta_2$ is a scaling constant, $p_{\boldsymbol{\phi}}(\mathbf{c}|\boldsymbol{\tau}^k) = B_{\boldsymbol{\phi}}$, and $\mathbf{c}=(c_t, \cdots, c_T)$ is the vector containing the class labels for each step of the trajectory. We can use the sampling procedure in Equation \eqref{eq:safe_planning_diffusion} to generate trajectories the are classified as safe by $B_{\boldsymbol{\phi}}$. Again if both initial and final positions are known, one can simply use \eqref{eq:safe_planning_diffusion} together with inpainting to generate safe trajectories connecting the two points. However, these trajectories may not be optimal in terms of reward accumulated, e.g. may be over-conservative or require high-control effort. 

Instead of guiding the planning towards optimality or safety, we propose to do both by employing both guides at the same time and introducing a new conditional sampling procedure:
\begin{equation}
    \boldsymbol{\tau}^{k-1} \sim \mathcal{N}(\boldsymbol{\mu}_{\boldsymbol{\theta}}+\boldsymbol{\Sigma}(\eta_1 \nabla f_1 + \eta_2\nabla f_2), \boldsymbol{\Sigma})
    \label{eq:optimalsafe_planning_diffusion}
\end{equation}
where $f_1=V_{\boldsymbol{\psi}}(\boldsymbol{\tau}^k)$ and $f_2=\log p_{\boldsymbol{\phi}}(\mathbf{c}|\boldsymbol{\tau}^k)$\footnote{We drop subscripts and superscripts for the sake of conciseness.}. Similarly to before, we can now generate optimal and safe trajectories by simply plugging \eqref{eq:optimalsafe_planning_diffusion} in Algorithm \ref{alg:optsafeplan}.

DDPMs are probabilistic models, thus the generated trajectories may differ every time we call Algorithm \ref{alg:optsafeplan}. Therefore, we do not simply generate a single trajectory but a batch of them (64 in our experiments), we then select one accordingly to a predefined criterium, and we apply the first action of the plan to the environment. Because our main concern is safety, we select the trajectory without collisions, but more advance criteria may be used, such as considering a weighted sum of value and collisions or even learning the weighting factor. However, we leave this interesting research direction to future work.

\subsection{Training Objectives}\label{subsec:training_obj}
To train our models and test their generalization capabilities in low-data regimes, we collect a dataset $\mathcal{D}$ with 300 trajectories of 100 steps each for a total of 30000 tuples $(\mathbf{s}_t^{\text{target}}, \mathbf{a}_t^{\text{target}}, \mathbf{s}_{t+1}^{\text{target}}, r_t^{\text{target}}, d_t^{\text{target}}, c_t^{\text{target}})$, where $r_t^{\text{target}}$ is the instantaneous reward, $d_t^{\text{target}}$ is a boolean flag indicating the end of the episode, and:
\begin{equation}
c_t^{\text{target}} = \begin{cases}
0 & \text{if   } \ \ \ h(\mathbf{s}_{t+1})-h(\mathbf{s}_{t})\geq -\alpha(h(\mathbf{s}_t))\\
1 &\text{otherwise}
\end{cases}
\end{equation}
is the label generated through the use of the CBF constraint in Equation \eqref{eq:CBFcondition}.
We use a random control policy to generate the training dataset. 

Similarly to \cite{Janner2022PlanningSynthesis}, instead of relying of the objective function in Equation \eqref{eq:diffusion_simple_objective}\footnote{For the example considered, we did not notice substantial difference between training the models to reconstruct the original signal or to reconstruct the noise.}, we train the diffusion model $D_{\boldsymbol{\theta}}$ using the mean-squared error loss between the generated trajectory $\boldsymbol{\tau}$ and the measured one $\boldsymbol{\tau}^{\text{target}}$:
\begin{equation}
    \mathcal{L}(\boldsymbol{\theta}) = \mathbb{E}_{\boldsymbol{\tau}_t, \boldsymbol{\tau}_t^{\text{target}} \sim \mathcal{D}}[||\boldsymbol{\tau}_t-\boldsymbol{\tau}_t^{\text{target}}||_2]
    \label{eq:d_loss}
\end{equation}
The value function model is instead trained to predict the expected return of a given trajectory, but again with mean-squared error loss:
\begin{equation}
    \mathcal{L}(\boldsymbol{\psi}) = \mathbb{E}_{\boldsymbol{\tau}_t, r_{t:H}} \sim \mathcal{D}[||v_t - v_t^{\text{target}}||_2]
     \label{eq:v_loss}
\end{equation}
where $v_t^{\text{target}}=\sum_{t=0}^{H-1}\gamma^t r_t^{\text{target}}$ with $H$ equal to the planning horizon and $\gamma \in [0, 1]$ is the discount factor. 
Eventually, the CBF classifier $B_{\boldsymbol{\phi}}$ is trained with the cross entropy loss:
\begin{equation}
     \mathcal{L}(\boldsymbol{\phi}) = \mathbb{E}_{\boldsymbol{\tau}_t, c_{t:T}\sim \mathcal{D}}[\sum_{t=0}^{T-1}-c_t^{\text{target}}\log(c_t)+(1-c_t^{\text{target}})\log(c_t)]
      \label{eq:b_loss}
\end{equation}

\section{Numerical Experiments}

\begin{figure}[h!]
    \centering
    \includegraphics[width=0.55\linewidth]{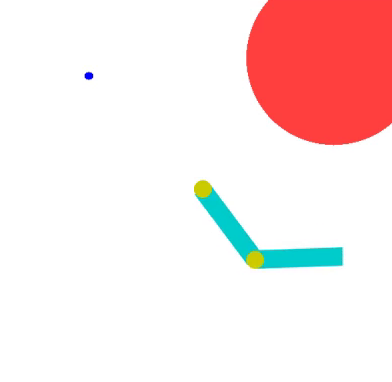}
    \caption{Openai Gym simulated environment. The blue dot represents the target location that the end-effector has to reach, while the red circle represent the unsafe region.}
    \label{fig:env}
\end{figure}

For our numerical experiments \footnote{The framework is implemented using PyTorch \cite{Paszke2019PyTorch:Library} and Openai Gym \cite{Brockman2016OpenAIGym}. The code can be found at: \href{https://github.com/nicob15/Trajectory-Generation-Control-and-Safety-with-Denoising-Diffusion-Probabilistic-Models}{{\fontfamily{cmtt}\selectfont github.com/nicob15}}.}, we simulate a two degrees of freedom, fully-actuated, planar manipulator in the task of reaching a randomly-sampled target from a randomly-sampled initial configuration of its angles and angular velocities. Additionally, we add a circular unsafe region, the end-effector must not enter into as shown in Figure \ref{fig:env}.

\begin{figure*}[ht!]
     \centering
     \begin{subfigure}{0.24\textwidth}
         \centering
         \includegraphics[width=\textwidth]{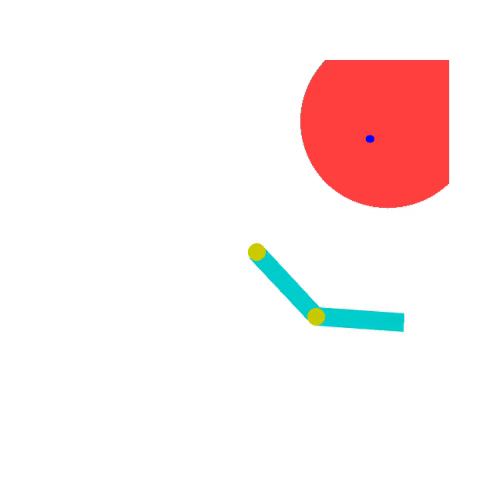}
         \caption{}
         \label{fig:t0}
     \end{subfigure}
     \hfill
     \begin{subfigure}{0.24\textwidth}
         \centering
         \includegraphics[width=\textwidth]{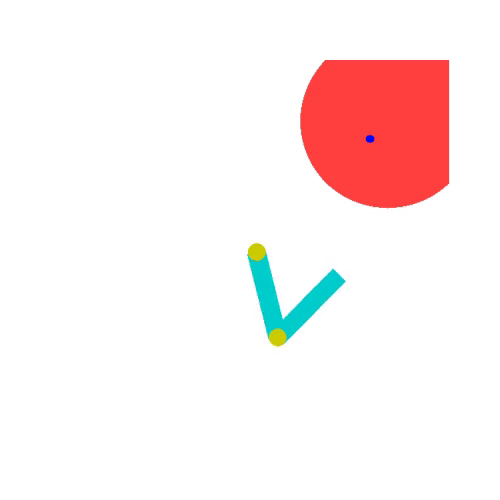}
         \caption{}
         \label{fig:t1}
     \end{subfigure}
     \hfill
     \begin{subfigure}{0.24\textwidth}
         \centering
         \includegraphics[width=\textwidth]{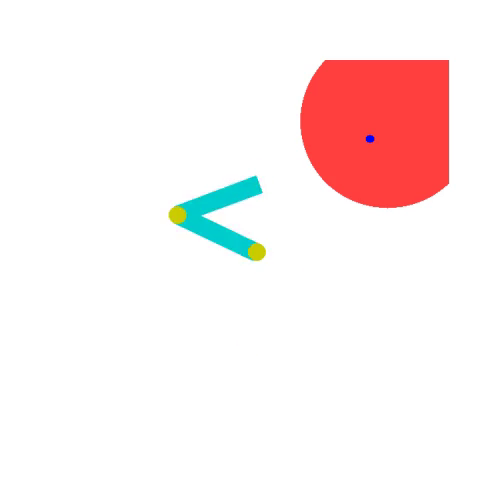}
         \caption{}
         \label{fig:t2}
     \end{subfigure}
          \begin{subfigure}{0.24\textwidth}
         \centering
         \includegraphics[width=\textwidth]{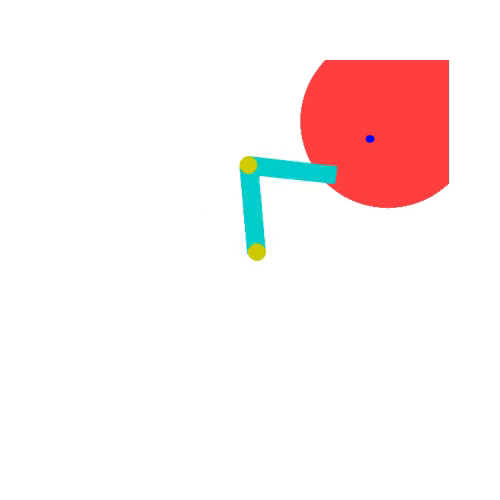}
         \caption{}
         \label{fig:t3}
     \end{subfigure}
          \centering
     \begin{subfigure}{0.24\textwidth}
         \centering
         \includegraphics[width=\textwidth]{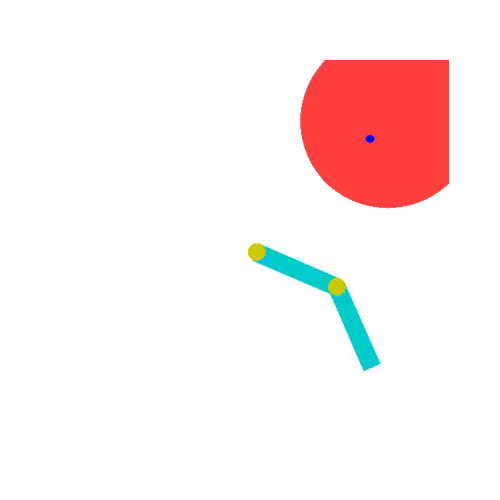}
         \caption{}
         \label{fig:t4}
     \end{subfigure}
     \hfill
     \begin{subfigure}{0.24\textwidth}
         \centering
         \includegraphics[width=\textwidth]{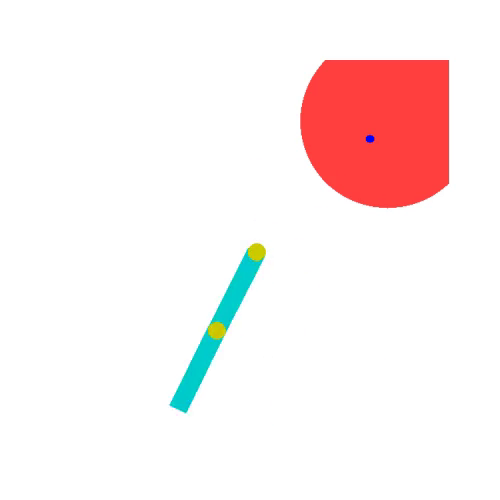}
         \caption{}
         \label{fig:t5}
     \end{subfigure}
     \hfill
     \begin{subfigure}{0.24\textwidth}
         \centering
         \includegraphics[width=\textwidth]{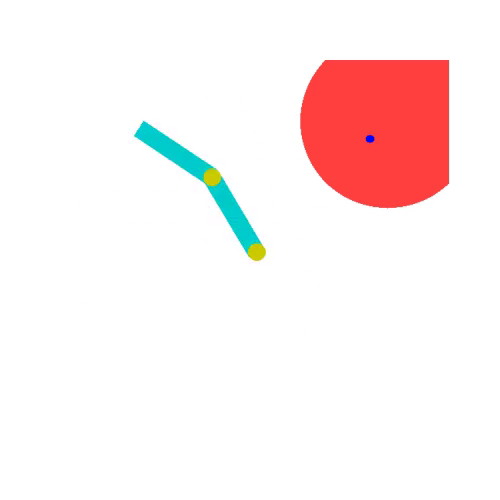}
         \caption{}
         \label{fig:t6}
     \end{subfigure}
          \begin{subfigure}{0.24\textwidth}
         \centering
         \includegraphics[width=\textwidth]{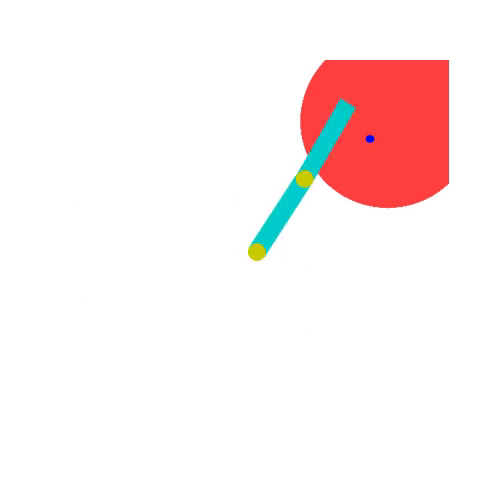}
         \caption{}
         \label{fig:t7}
     \end{subfigure}
        \caption{Trajectory generated by the diffusion planner with value-function guide.}
        \label{fig:traj_value_only}
\end{figure*}

The state is composed of the joint angles $\alpha_1$ and $\alpha_2$, expressed with their sine and cosine, their angular velocity $\Dot{\alpha}_1$ and $\Dot{\alpha}_2$, and the target position $(x_{tg}, y_{tg})$ for a total dimension of 8. The actions $a_1, a_2$ are the torques at the two joints $\in [-1.0, 1.0]$. Because the goal of the agent is to reach a target position starting from any initial configuration, we chose as the reward function the negative of the Euclidean distance from the end-effector to the target:
\begin{equation}
    R(\mathbf{s}) = -d(x_{ee}, y_{ee}, x_{tg}, y_{tg}) 
\end{equation}
with $x_{ee}, y_{ee}$ indicating the coordinates of the end-effector, and $d$ indicating the Euclidean distance. This reward function, when maximized, corresponds to the end-effector reaching the target.

To assess the performance of this framework in low data regimes, we first train offline  $D_{\boldsymbol{\theta}}$, $V_{\boldsymbol{\psi}}$, and $B_{\boldsymbol{\phi}}$ using a small dataset of randomly generated trajectory (as described in Section \ref{subsec:training_obj}) with loss functions \eqref{eq:d_loss}-\eqref{eq:b_loss} respectively and then we evaluate the policy generated using the guided planning generated Algorithm \ref{alg:optsafeplan} using only the value function guide on a set of 100 randomly-generated target. We record success rate\footnote{We consider success when the end-effect reaches the target, with a predefined tolerance $\epsilon$, withing 100 steps, and unsuccess otherwise.}, average rewards, average number of steps, and standard deviation of them. We compare with soft-actor critic (SAC) \cite{Haarnoja2018SoftApplications} trained for 100K, 200K, and 300K iteration for 300 episodes of 100 steps each for the sake of a fair comparison. We report the results in Table \ref{tab:quantitative_results}. The complete list of hyperparameters and an ablation study on the guide scale $\eta_1$ can be find respectively in Appendix \ref{app:hyper} and \ref{app:ablation}.

Secondly, we want to test whether we can generate optimal and safe trajectories using Algorithm \ref{alg:optsafeplan} by using both the value and the safety-classifier guide accordingly to Equation \eqref{eq:optimalsafe_planning_diffusion}.

\section{Results}
From Table \ref{tab:quantitative_results}, we can notice that the proposed offline and model-based approach in low-data regimes is able to generalize better that model-free SAC in terms of success rate and average number of steps to reach the target, but not in term of average reward. The small difference is likely due to the different dataset used for training the two methods. Despite the two methods are trained on the same amount of data, the diffusion models are trained completely offline on randomly generated trajectories, while  SAC is trained online. Therefore, SAC is able to experience more high-reward trajectories in later episodes. 
\begin{table}[h!]
\tiny
    \centering
    \begin{tabular}{||c|c|c|c||}
    \hline
    \hline
     Method & Succ. Rate & Avg. Rew. & Avg. Steps \\
    \hline
    \hline
     $D_{\boldsymbol{\theta}} + V_{\boldsymbol{\psi}}$ (200K it.) & $0.65$ & $-104.24 \pm 78.06$ & $53.42 \pm 38.42$ \\
     \hline
     SAC (300K it.) & $0.57$  & $-98.67\pm 74.44$ & $55.97 \pm 40.89$ \\
     \hline
     SAC (200K it.)   & $0.56$  & $-102.3\pm 68.93 $ & $ 59.01\pm 38.09 $ \\
     \hline
     SAC (100K it.)   & $0.55$  & $-107.07\pm 69.18$ & $ 62.71\pm 38.59$ \\
     \hline
    \end{tabular}
    \caption{Comparison of the proposed offline, model-based approach trained for 200K iteration and online, model-free SAC trained for 300K, 200K, 100K iterations on a set of 100 randomly-generated target and initial conditions.}
    \label{tab:quantitative_results}
\end{table}

\begin{figure*}[h!]
     \centering
     \begin{subfigure}{0.24\textwidth}
         \centering
         \includegraphics[width=\textwidth]{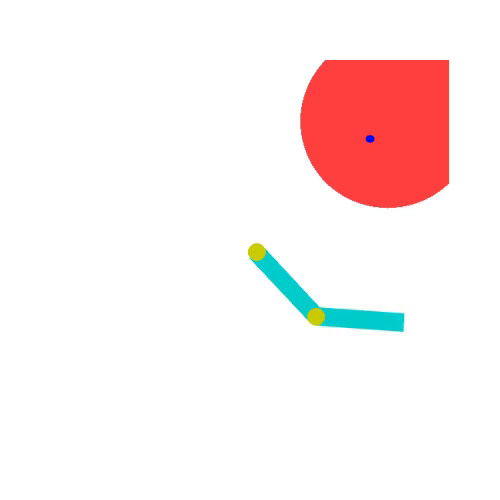}
         \caption{}
         \label{fig:cbft0}
     \end{subfigure}
     \hfill
     \begin{subfigure}{0.24\textwidth}
         \centering
         \includegraphics[width=\textwidth]{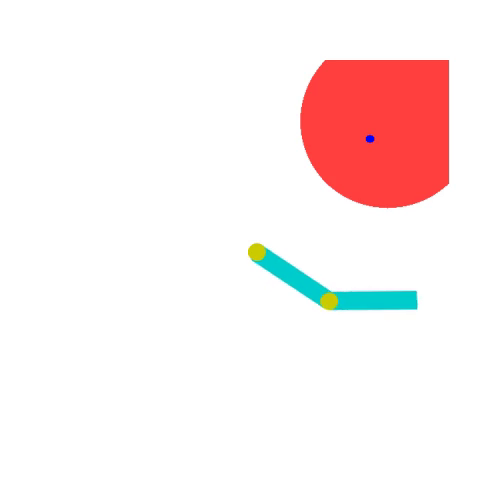}
         \caption{}
         \label{fig:cbft1}
     \end{subfigure}
     \hfill
     \begin{subfigure}{0.24\textwidth}
         \centering
         \includegraphics[width=\textwidth]{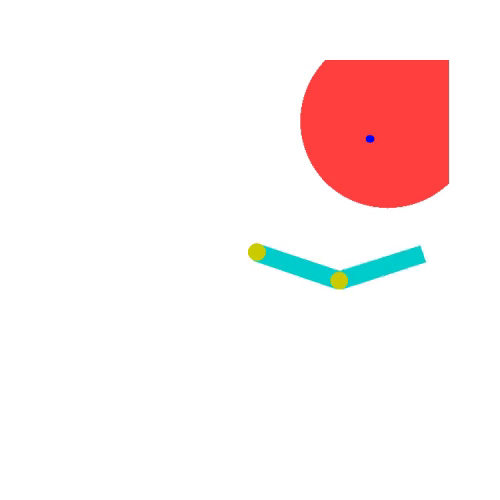}
         \caption{}
         \label{fig:cbft2}
     \end{subfigure}
          \begin{subfigure}{0.24\textwidth}
         \centering
         \includegraphics[width=\textwidth]{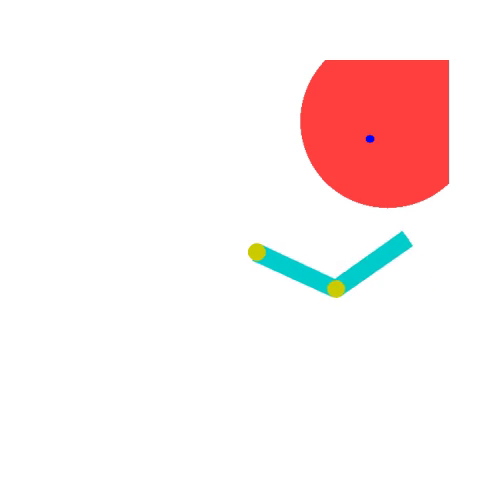}
         \caption{}
         \label{fig:cbft3}
     \end{subfigure}
          \centering
     \begin{subfigure}{0.24\textwidth}
         \centering
         \includegraphics[width=\textwidth]{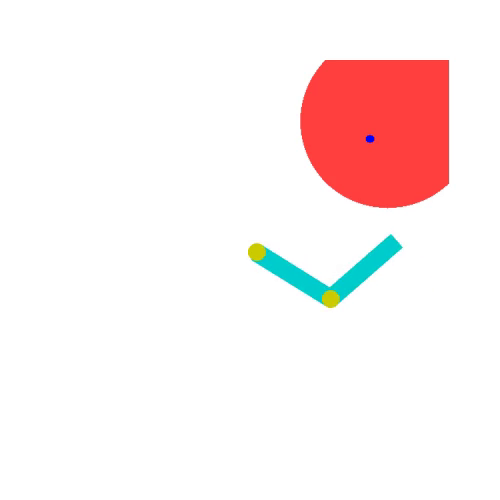}
         \caption{}
         \label{fig:cbft4}
     \end{subfigure}
     \hfill
     \begin{subfigure}{0.24\textwidth}
         \centering
         \includegraphics[width=\textwidth]{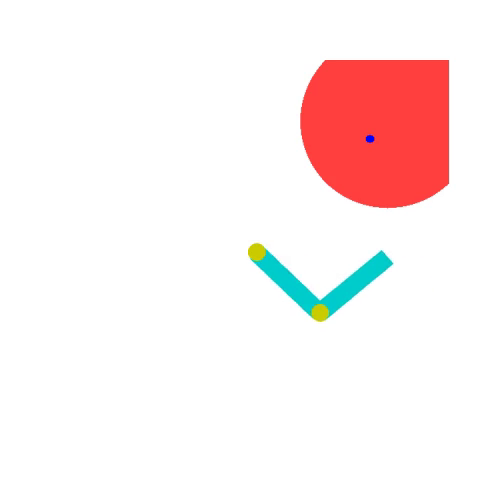}
         \caption{}
         \label{fig:cbft5}
     \end{subfigure}
     \hfill
     \begin{subfigure}{0.24\textwidth}
         \centering
         \includegraphics[width=\textwidth]{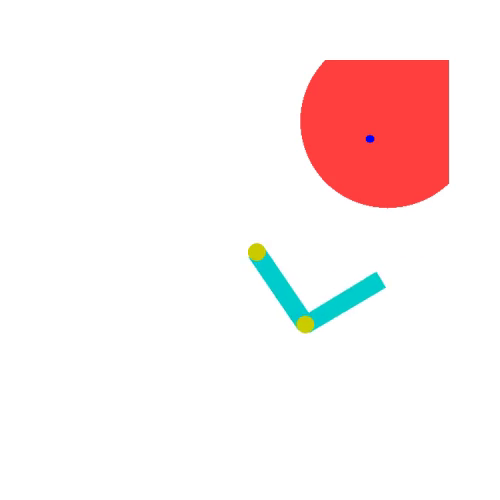}
         \caption{}
         \label{fig:cbft6}
     \end{subfigure}
          \begin{subfigure}{0.24\textwidth}
         \centering
         \includegraphics[width=\textwidth]{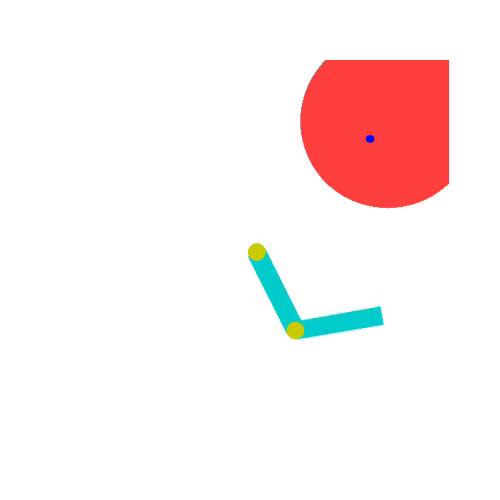}
         \caption{}
         \label{fig:cbft7}
     \end{subfigure}
        \caption{Trajectory generated by the diffusion planner with value-function and safety-classifier guides.}
        \label{fig:traj_value_cbf}
\end{figure*}

In Figure \ref{fig:traj_value_only} and \ref{fig:traj_value_cbf}, we show the trajectories generated by the DDPMs with value guide (Equation \eqref{eq:planning_diffusion}) and with the proposed value and safety-classifier guide (Equation \eqref{eq:optimalsafe_planning_diffusion}). It is worth mentioning that the dataset used to train the models does not contain targets in the unsafe region, as in this case. Therefore, this is a good example of generalization capabilities of the diffusion models even in low-data regimes. While the diffusion planner can reach the target (see Figure \ref{fig:traj_value_only}), without any safety constraint it naturally ends up in the unsafe region multiple times. On the other side, the diffuser planner employing the conditional sampling strategy in \eqref{eq:optimalsafe_planning_diffusion} does not violate the safery region while close to the (unreachable) target. Additional results can be found in Appendix \ref{app:additional_experiments}.

\section{Conclusion}
In this work, we studied the problem of safety-critical optimal control using data-driven generative models, i.e. DDPMs. In particular, we transform the problem of learning an optimal and safe policy into the problem of conditional sampling of trajectories, i.e. state-action pairs over an horizon $H$, either with high value and safe using a value function model and a safety (CBF-inspired) classifier as guides. We show that the method is not only more sample efficient that model-free RL in low-data regimes, but it can also be used to generate trajectories avoiding the unsafe regions. Additionally, we decoupled trajectory generation, value estimation, and safety prediction making by using three independent DDPMs, making the framework extremely flexible when dealing with changes in the environment, in the task, or in the safety regions.


\bibliography{references}
\bibliographystyle{icml2023}

\newpage
\appendix
\onecolumn

\section{Architecture}\label{app:Architecture}

We use a similar architecture to \cite{Janner2022PlanningSynthesis} for the diffusion models $D_{\boldsymbol{\theta}}$, value function $V_{\boldsymbol{\psi}}$, safety classifier $B_{\boldsymbol{\phi}}$. The three models use a U-net with 6 residual blocks with linear attention. Each block is composed of two 1D temporal convolutions, group normalization \cite{Wu2020GroupNormalization} and Mish activation \cite{Misra2019Mish:Function}. The timesteps of the diffusion process are encoded via a fully-connected layers in the first temporal convolution of each block. The value function $V_{\boldsymbol{\psi}}$ and the barrier classifier $B_{\boldsymbol{\phi}}$ additionally have a final fully-connected layer to output a single scalar and a vector of dimension equal to the number classes (2) $\times H$ with $H$ the planning horizon.

\section{Hyperparameters}\label{app:hyper}
In Table \ref{tab:hyper}, the hyperparameters of the experiments are reported.
\begin{table}[h!]
    \centering
\begin{tabular}{||c | c ||} 
 \hline
 Hyperparameter & Value \\ [0.5ex] 
 \hline\hline
 optimizer & AdamW \cite{Loshchilov2019DecoupledRegularization} \\
 \hline
 learning rate & $2e-4$ \\
 \hline
 batch size & $32$ \\
 \hline
 training dataset size & $30K$ \\
 \hline
 testing dataset size & $3K$ \\
 \hline
 training epochs & $200$ \\
 \hline
 training iterations & $[\mathbf{200K}, 300K]$ \\
 \hline
 num. diffusion steps & $50$ \\ 
 \hline
 planning horizon & $16$ \\
  \hline
CBF constant $\lambda$ &$ 0.99$ \\
 \hline
 value function discount $\gamma$ &$ 0.997$ \\
 \hline
 value guide scale $\eta_1$ &$ [0.1, 0.01, \mathbf{0.001}, 0.0005, 0.0001]$ \\
  \hline
 classifier guide scale $\eta_2$ & $[0.1, 1.0, \mathbf{5.0}, 10.0]$ \\
 \hline
 tolerance $\epsilon$ & $0.3$ \\
 \hline
 state dim. & $8$  \\
 \hline
 action dim. & $2$ \\ [0.5ex] 
 \hline
\end{tabular}
    \caption{DDPMs hyperparameters.}
    \label{tab:hyper}
\end{table}

The SAC hyperparameters used are presented in Table \ref{tab:hyper_sac}.
\begin{table}[h!]
    \centering
\begin{tabular}{||c | c ||} 
 \hline
 Hyperparameter & Value \\ [0.5ex] 
 \hline\hline
 optimizer & Adam \cite{Kingma2015Adam:Optimization}\\
 \hline
 learning rate & $3e-4$\\
 \hline
 batch size & $3$2 \\
 \hline
 num. training episodes & $300$ \\
 \hline
 num. training steps & $100$ \\
 \hline
 training dataset size & $30K$ \\
  \hline
 training iterations & $[100K, 200K, 300K]$ \\
 \hline
 value function discount $\gamma$ & $[\mathbf{0.99}, 0.997]$ \\
 \hline
 num. hidden layers & $2$ \\
 \hline
 hidden neurons & $256$ \\
 \hline
 hidden layers activation & ReLU \\
 \hline
 output activation & Tanh \\
  \hline
 tolerance $\epsilon$ & $0.3$ \\
 \hline
 state dim. & $8$  \\
 \hline
 action dim. & $2$ \\ 
 \hline
 num. different seeds & $3$ \\ [0.5ex] 
 \hline
\end{tabular}
    \caption{SAC hyperparameters.}
    \label{tab:hyper_sac}
\end{table}

\section{Ablation Study}\label{app:ablation}

\subsection{Guide Scale Selection}
To select the value function guide scale hyperparameter $\eta_1$, we performed a grid search on a set of 100 randomly-generated targets and initial conditions using models trained $200K$ iterations. The results are reported in Table \ref{tab:guide_scale}
\begin{center}
\begin{tabular}{||c | c| c | c ||} 
 \hline
 Guide Scale & Succ. Rate & Avg. Rew. & Avg. Steps \\ [0.5ex] 
 \hline\hline
$0.1$ & $0.54$ & $-123.81 \pm 84.6$ & $61.75 \pm 39.83$ \\
 \hline
 $0.01$ & $0.47$ & $-122.49 \pm 88.76$ & $63.11 \pm 41.18$ \\
 \hline
 $0.001$ & $0.65$ & $-104.24 \pm 78.06$ & $53.42 \pm 38.42$ \\
  \hline
 $0.0005$ & $0.58$ & $-104.44 \pm 85.82$ & $53.76 \pm 41.94$ \\
  \hline
 $0.0001$ & $0.51$ & $-125.63 \pm 72.31$ & $66.34 \pm 37.14$ \\[1ex] 
 \hline
\end{tabular}
\label{tab:guide_scale}
\end{center}
We also performed similar experiments on a model trained for $300K$ iterations, but we did not notice substantial change in the performance.

\section{Additional Experiments}\label{app:additional_experiments}
Figure \ref{fig:traj_value_only} and \ref{fig:traj_value_cbf} show snapshots of a trajectory generated by following the plan generated by Algorithm \ref{alg:optsafeplan} with value guide (see Equation \eqref{eq:planning_diffusion}) and with value and safety-classifier guide (see Equation \eqref{eq:optimalsafe_planning_diffusion}). Analogously in this section, we show additional results achieved by the two different planning strategies. The code of our experiments can be found at: \href{https://github.com/nicob15/Trajectory-Generation-Control-and-Safety-with-Denoising-Diffusion-Probabilistic-Models}{{\fontfamily{cmtt}\selectfont github.com/nicob15}}.

\begin{figure*}[h!]
     \centering
     \begin{subfigure}{0.24\textwidth}
         \centering
         \includegraphics[width=\textwidth]{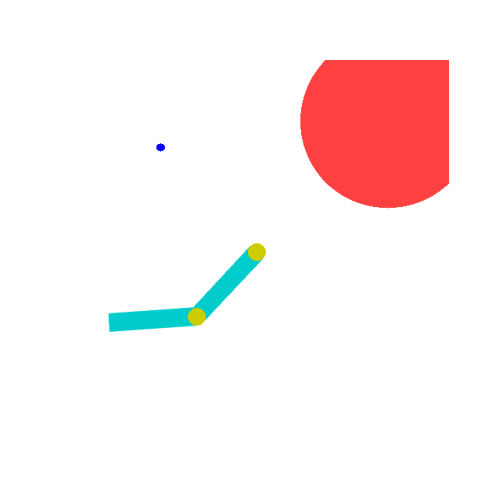}
         \caption{}
     \end{subfigure}
     \hfill
               \begin{subfigure}{0.24\textwidth}
         \centering
         \includegraphics[width=\textwidth]{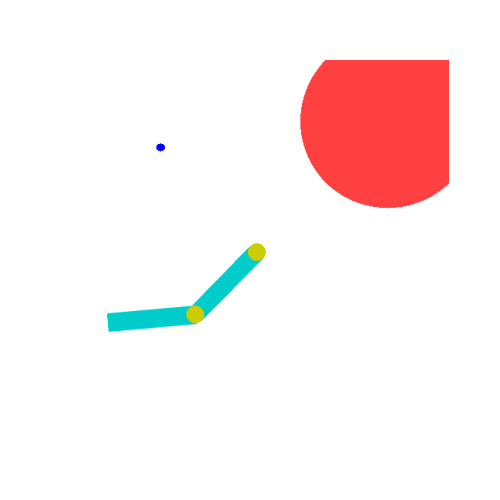}
         \caption{}
     \end{subfigure}
     \hfill
     \begin{subfigure}{0.24\textwidth}
         \centering
         \includegraphics[width=\textwidth]{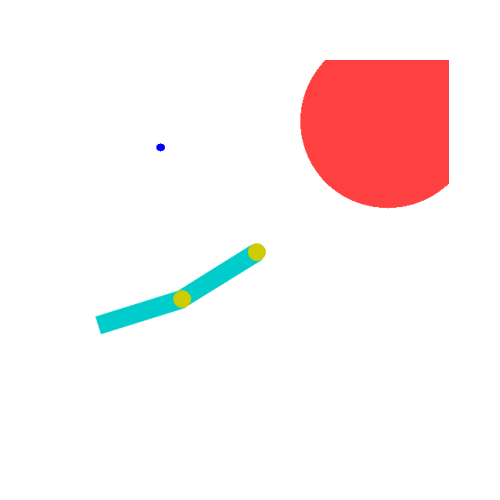}
         \caption{}
     \end{subfigure}
     \hfill
     \begin{subfigure}{0.24\textwidth}
         \centering
         \includegraphics[width=\textwidth]{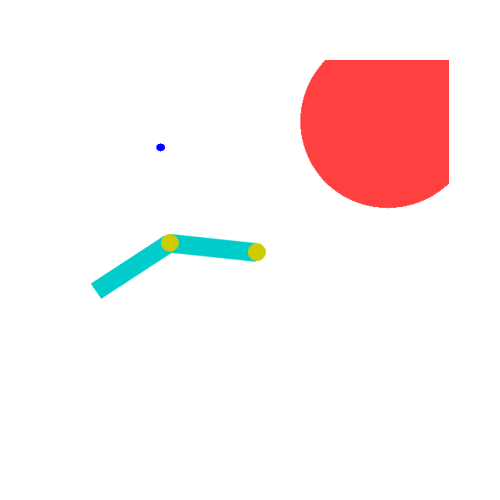}
         \caption{}
     \end{subfigure}
          \begin{subfigure}{0.24\textwidth}
         \centering
         \includegraphics[width=\textwidth]{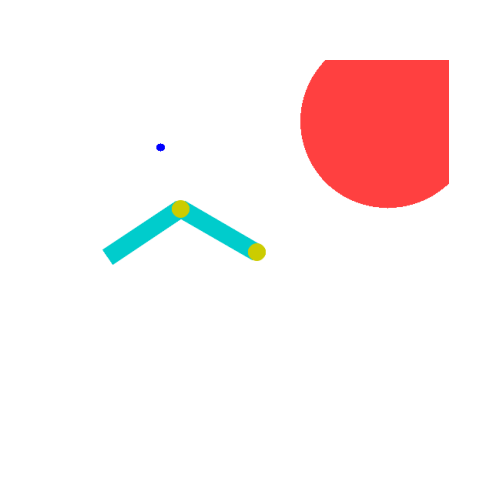}
         \caption{}
     \end{subfigure}
          \centering
     \begin{subfigure}{0.24\textwidth}
         \centering
         \includegraphics[width=\textwidth]{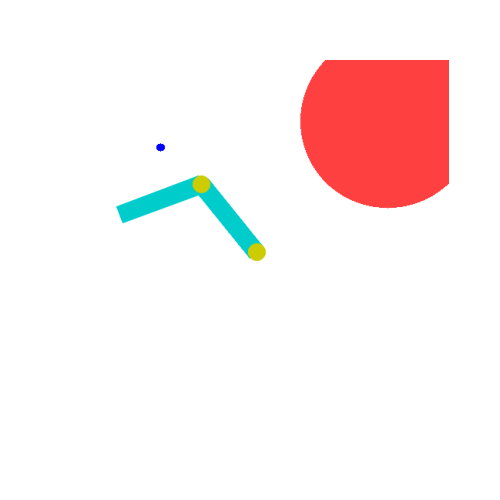}
         \caption{}
     \end{subfigure}
     \hfill
     \begin{subfigure}{0.24\textwidth}
         \centering
         \includegraphics[width=\textwidth]{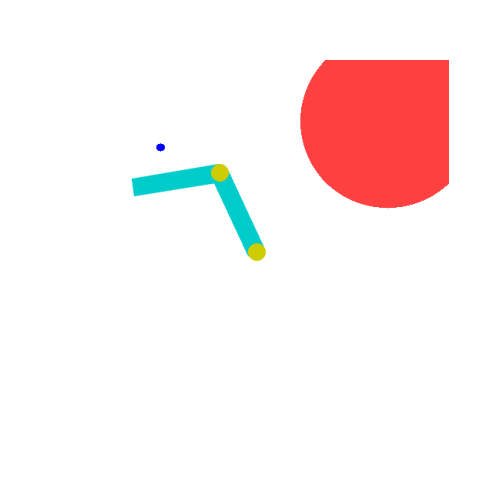}
         \caption{}
     \end{subfigure}
     \hfill
     \begin{subfigure}{0.24\textwidth}
         \centering
         \includegraphics[width=\textwidth]{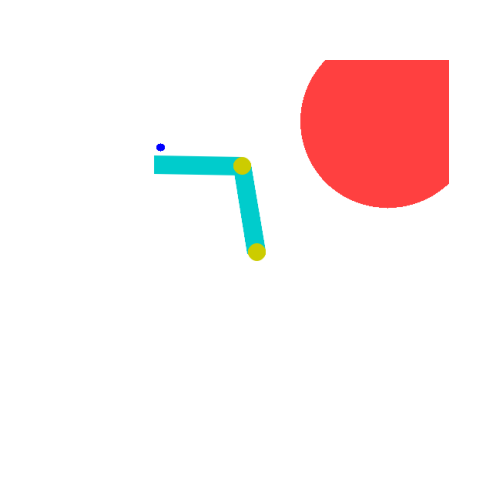}
         \caption{}
     \end{subfigure}
        \caption{Trajectory generated by the diffusion planner with value-function guide.}
        \label{fig:traj_value_only_extra1}
\end{figure*}

\begin{figure*}[h!]
     \centering
     \begin{subfigure}{0.24\textwidth}
         \centering
         \includegraphics[width=\textwidth]{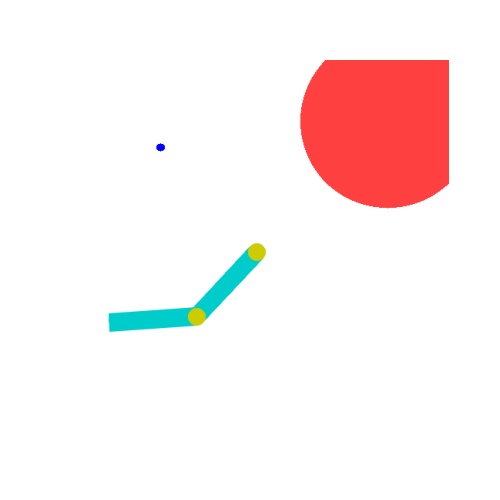}
         \caption{}
     \end{subfigure}
     \hfill
     \begin{subfigure}{0.24\textwidth}
         \centering
         \includegraphics[width=\textwidth]{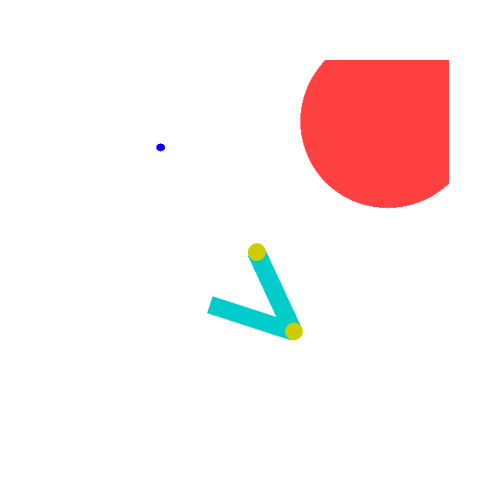}
         \caption{}
     \end{subfigure}
     \hfill
     \begin{subfigure}{0.24\textwidth}
         \centering
         \includegraphics[width=\textwidth]{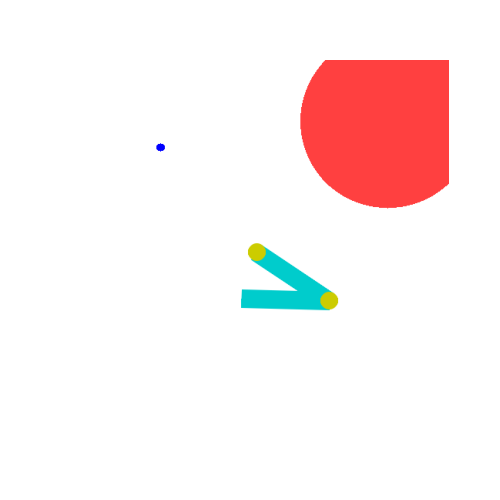}
         \caption{}
     \end{subfigure}
          \begin{subfigure}{0.24\textwidth}
         \centering
         \includegraphics[width=\textwidth]{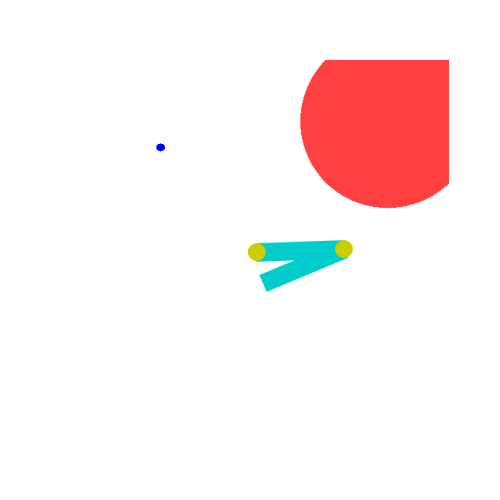}
         \caption{}
     \end{subfigure}
          \centering
     \begin{subfigure}{0.24\textwidth}
         \centering
         \includegraphics[width=\textwidth]{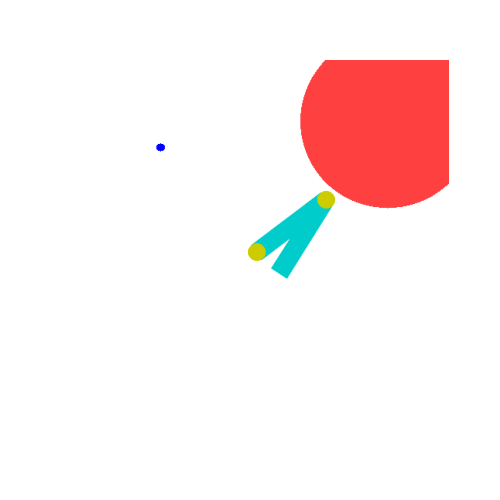}
         \caption{}
     \end{subfigure}
     \hfill
     \begin{subfigure}{0.24\textwidth}
         \centering
         \includegraphics[width=\textwidth]{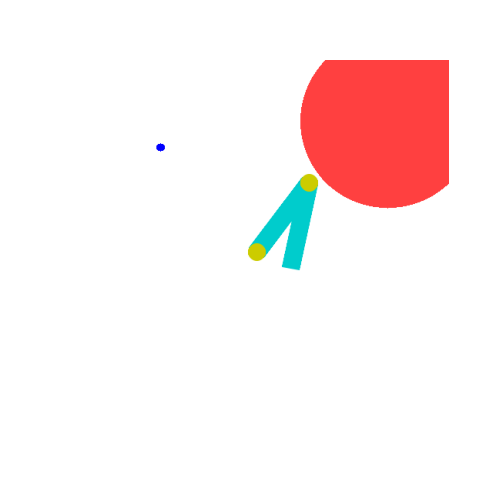}
         \caption{}
     \end{subfigure}
     \hfill
     \begin{subfigure}{0.24\textwidth}
         \centering
         \includegraphics[width=\textwidth]{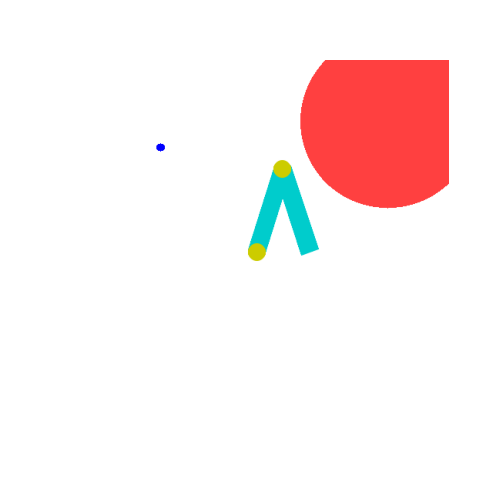}
         \caption{}
     \end{subfigure}
     \hfill
          \begin{subfigure}{0.24\textwidth}
         \centering
         \includegraphics[width=\textwidth]{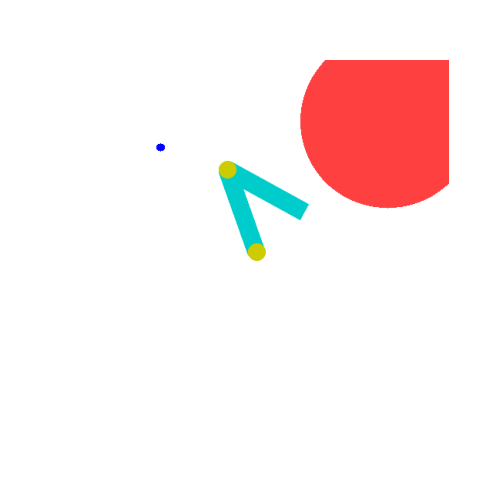}
         \caption{}
     \end{subfigure}
     \hfill
          \begin{subfigure}{0.24\textwidth}
         \centering
         \includegraphics[width=\textwidth]{Figures/target_2/value_cbf_guide/snapshots_39.png}
         \caption{}
     \end{subfigure}
          \hfill
          \begin{subfigure}{0.24\textwidth}
         \centering
         \includegraphics[width=\textwidth]{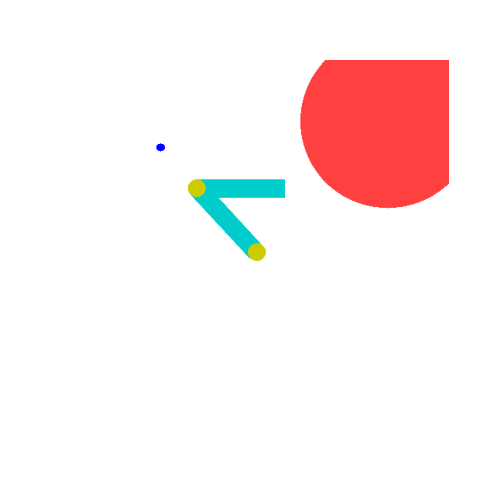}
         \caption{}
     \end{subfigure}
               \hfill
          \begin{subfigure}{0.24\textwidth}
         \centering
         \includegraphics[width=\textwidth]{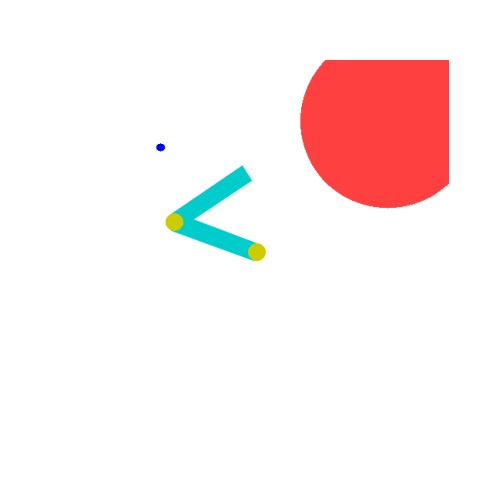}
         \caption{}
     \end{subfigure}
               \hfill
          \begin{subfigure}{0.24\textwidth}
         \centering
         \includegraphics[width=\textwidth]{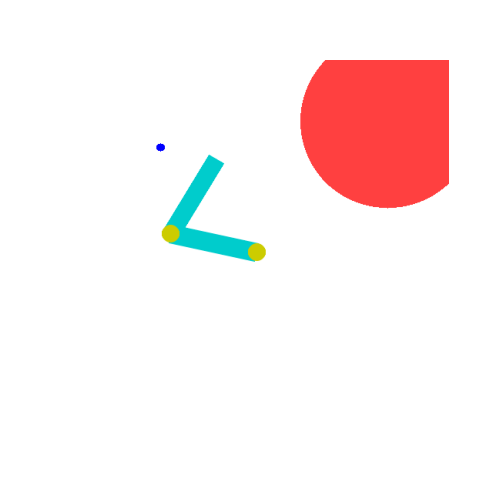}
         \caption{}
     \end{subfigure}
               \hfill
          \begin{subfigure}{0.24\textwidth}
         \centering
         \includegraphics[width=\textwidth]{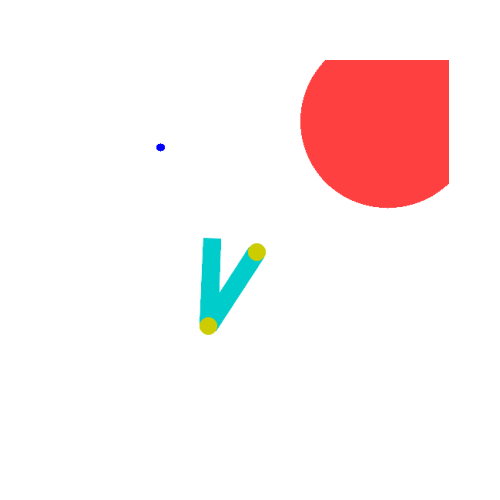}
         \caption{}
     \end{subfigure}
                    \hfill
          \begin{subfigure}{0.24\textwidth}
         \centering
         \includegraphics[width=\textwidth]{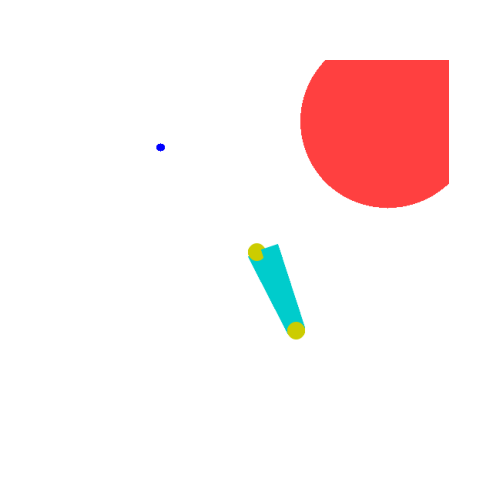}
         \caption{}
     \end{subfigure}
                    \hfill
          \begin{subfigure}{0.24\textwidth}
         \centering
         \includegraphics[width=\textwidth]{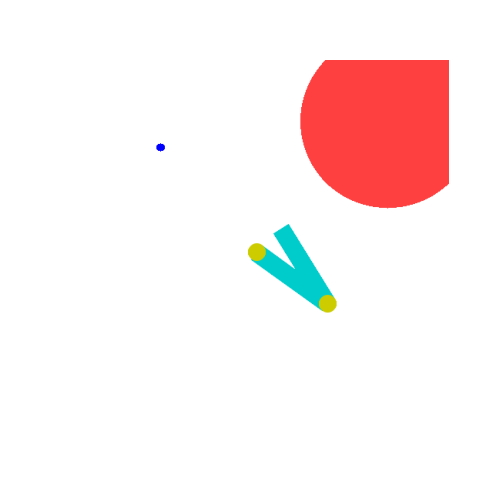}
         \caption{}
     \end{subfigure}
                    \hfill
          \begin{subfigure}{0.24\textwidth}
         \centering
         \includegraphics[width=\textwidth]{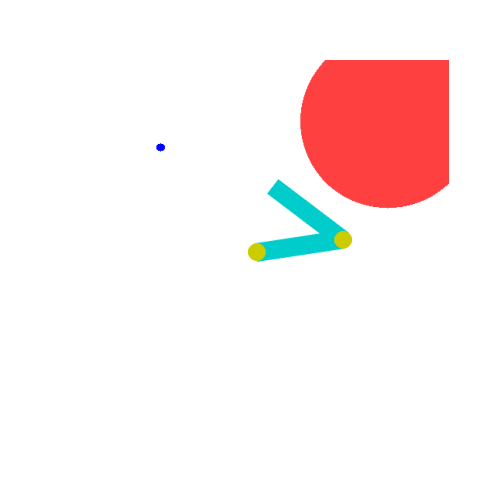}
         \caption{}
     \end{subfigure}
          \begin{subfigure}{0.24\textwidth}
         \centering
         \includegraphics[width=\textwidth]{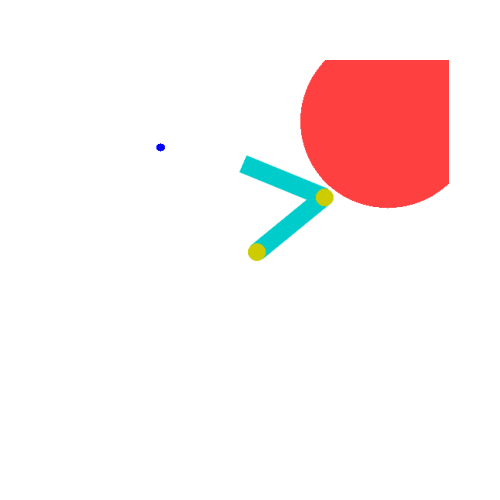}
         \caption{}
     \end{subfigure} 
          \begin{subfigure}{0.24\textwidth}
         \centering
         \includegraphics[width=\textwidth]{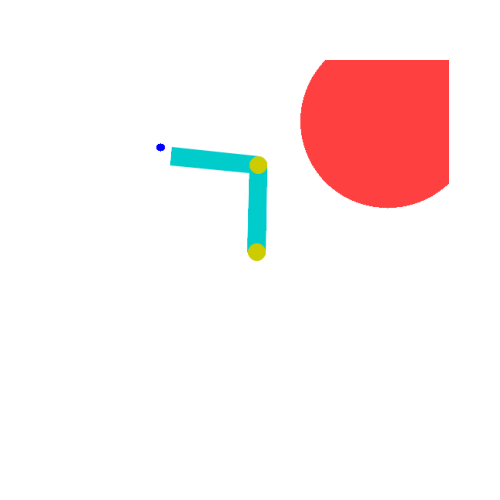}
         \caption{}
     \end{subfigure}
        \caption{Trajectory generated by the diffusion planner with value-function and safety-classifier guides.}
        \label{fig:traj_value_cbf_extra1}
\end{figure*}


\begin{figure*}[h!]
     \centering
     \begin{subfigure}{0.24\textwidth}
         \centering
         \includegraphics[width=\textwidth]{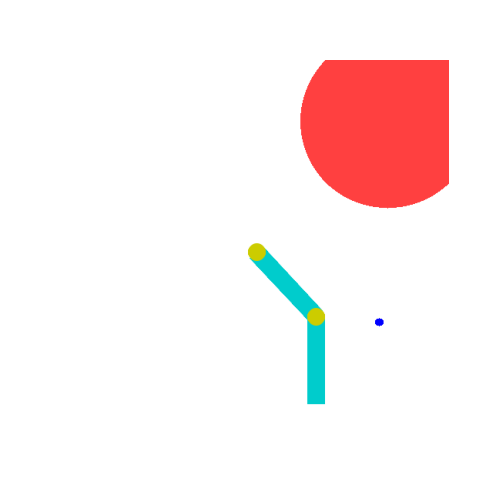}
         \caption{}
     \end{subfigure}
     \hfill
               \begin{subfigure}{0.24\textwidth}
         \centering
         \includegraphics[width=\textwidth]{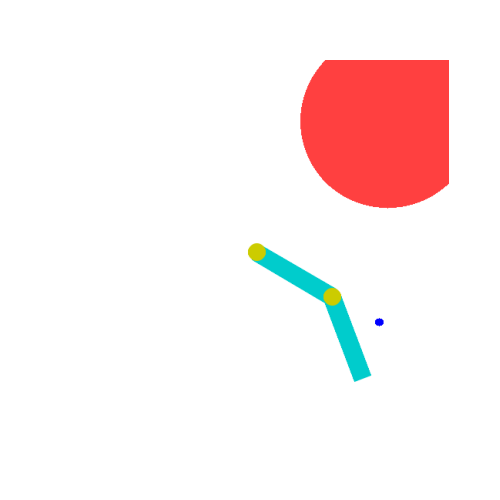}
         \caption{}
     \end{subfigure}
     \hfill
     \begin{subfigure}{0.24\textwidth}
         \centering
         \includegraphics[width=\textwidth]{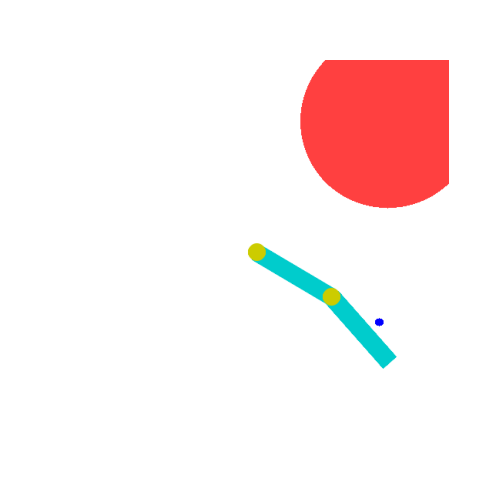}
         \caption{}
     \end{subfigure}
     \hfill
     \begin{subfigure}{0.24\textwidth}
         \centering
         \includegraphics[width=\textwidth]{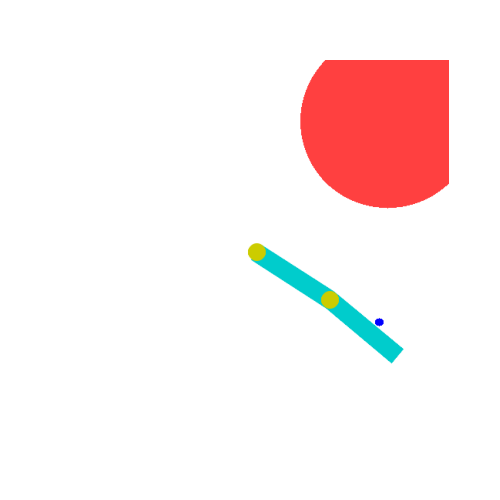}
         \caption{}
     \end{subfigure}    
     \begin{subfigure}{0.24\textwidth}
         \centering
         \includegraphics[width=\textwidth]{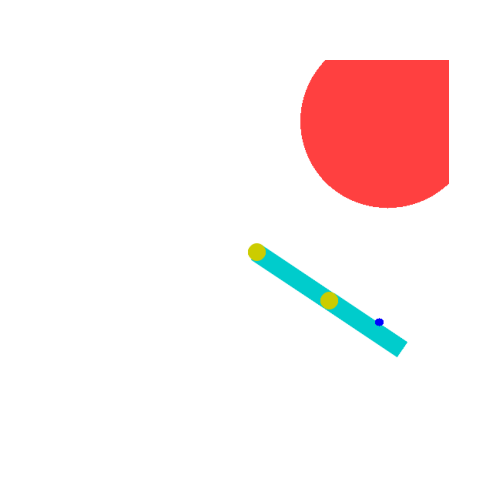}
         \caption{}
     \end{subfigure}
     \begin{subfigure}{0.24\textwidth}
         \centering
         \includegraphics[width=\textwidth]{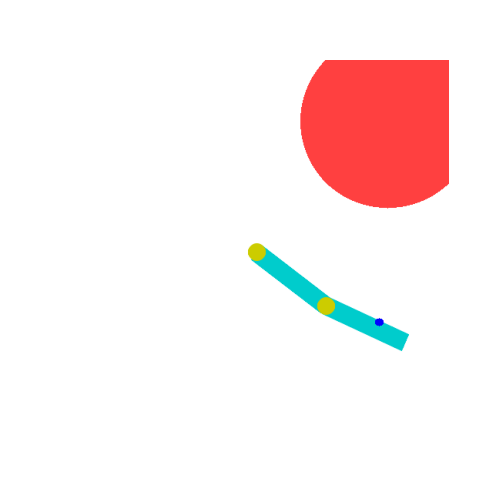}
         \caption{}
     \end{subfigure}   
     \begin{subfigure}{0.24\textwidth}
         \centering
         \includegraphics[width=\textwidth]{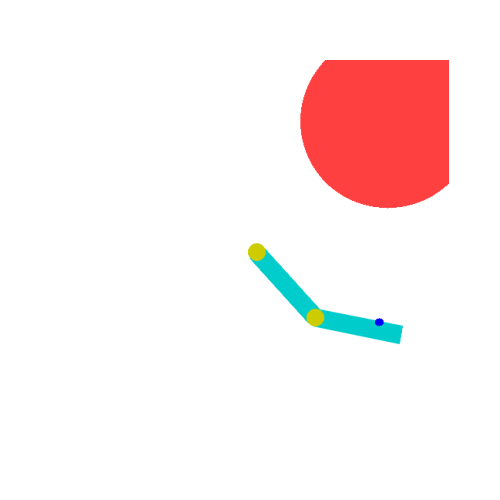}
         \caption{}
     \end{subfigure}
        \caption{Trajectory generated by the diffusion planner with value-function guide.}
        \label{fig:traj_value_only_extra2}
\end{figure*}

\begin{figure*}[h!]
     \centering
     \begin{subfigure}{0.24\textwidth}
         \centering
         \includegraphics[width=\textwidth]{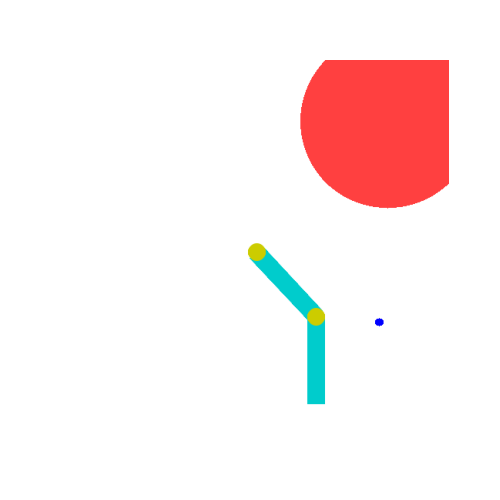}
         \caption{}
     \end{subfigure}
     \hfill
     \begin{subfigure}{0.24\textwidth}
         \centering
         \includegraphics[width=\textwidth]{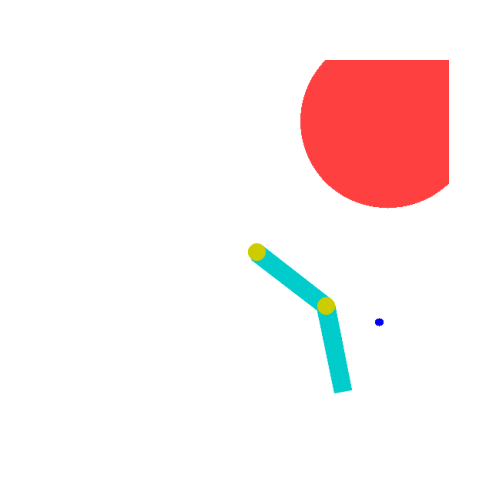}
         \caption{}
     \end{subfigure}
     \hfill
     \begin{subfigure}{0.24\textwidth}
         \centering
         \includegraphics[width=\textwidth]{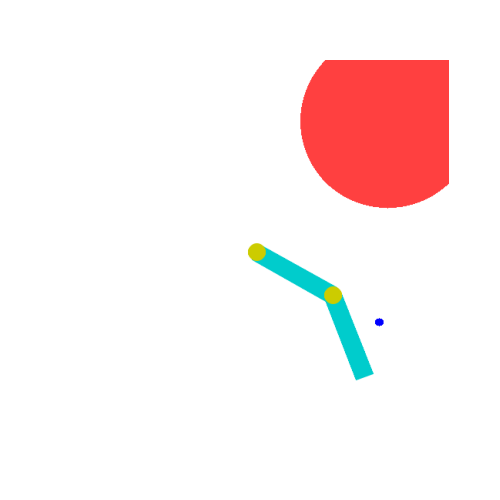}
         \caption{}
     \end{subfigure}
          \hfill
     \begin{subfigure}{0.24\textwidth}
         \centering
         \includegraphics[width=\textwidth]{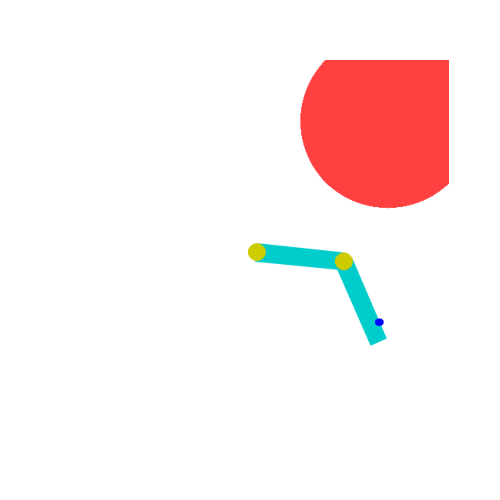}
         \caption{}
     \end{subfigure}
        \caption{Trajectory generated by the diffusion planner with value-function and safety-classifier guides.}
        \label{fig:traj_value_cbf_extra2}
\end{figure*}


\begin{figure*}[h!]
     \centering
     \begin{subfigure}{0.24\textwidth}
         \centering
         \includegraphics[width=\textwidth]{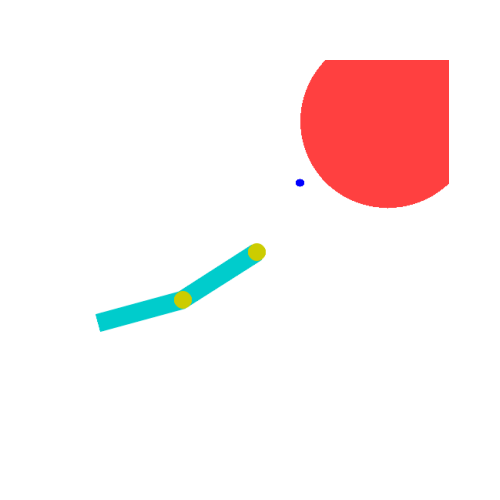}
         \caption{}
     \end{subfigure}
     \hfill
               \begin{subfigure}{0.24\textwidth}
         \centering
         \includegraphics[width=\textwidth]{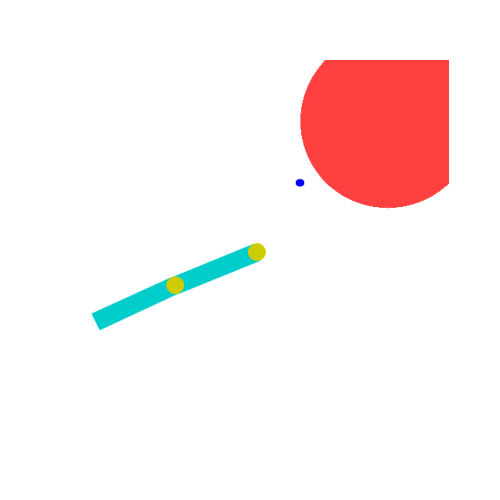}
         \caption{}
     \end{subfigure}
     \hfill
     \begin{subfigure}{0.24\textwidth}
         \centering
         \includegraphics[width=\textwidth]{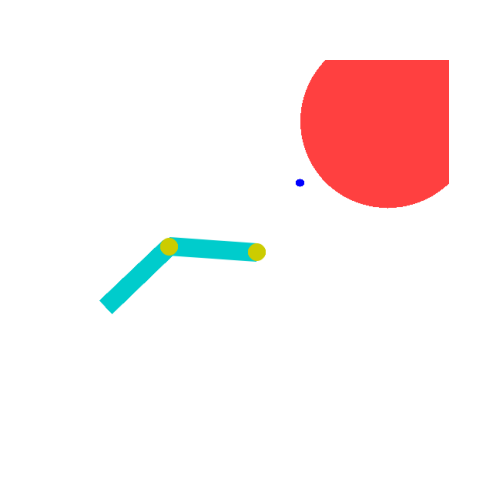}
         \caption{}
     \end{subfigure}
     \hfill
     \begin{subfigure}{0.24\textwidth}
         \centering
         \includegraphics[width=\textwidth]{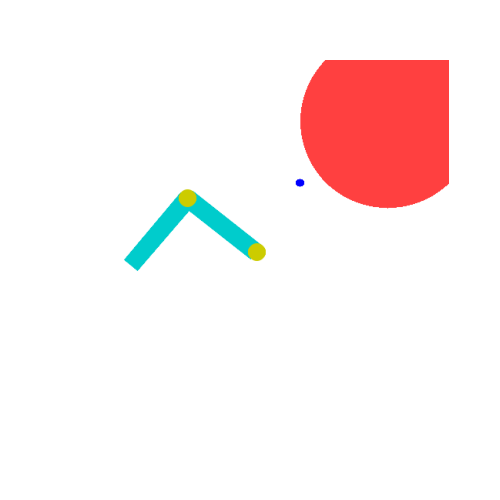}
         \caption{}
     \end{subfigure}
          \hfill
     \begin{subfigure}{0.24\textwidth}
         \centering
         \includegraphics[width=\textwidth]{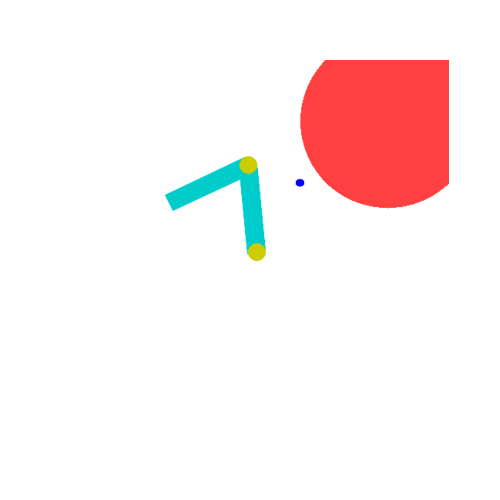}
         \caption{}
     \end{subfigure}     
     \hfill
     \begin{subfigure}{0.24\textwidth}
         \centering
         \includegraphics[width=\textwidth]{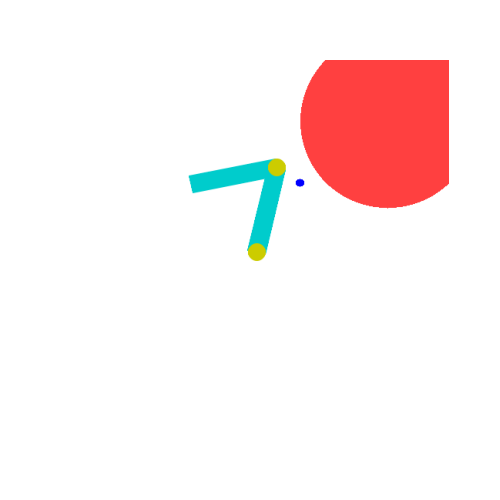}
         \caption{}
     \end{subfigure}
          \hfill
     \begin{subfigure}{0.24\textwidth}
         \centering
         \includegraphics[width=\textwidth]{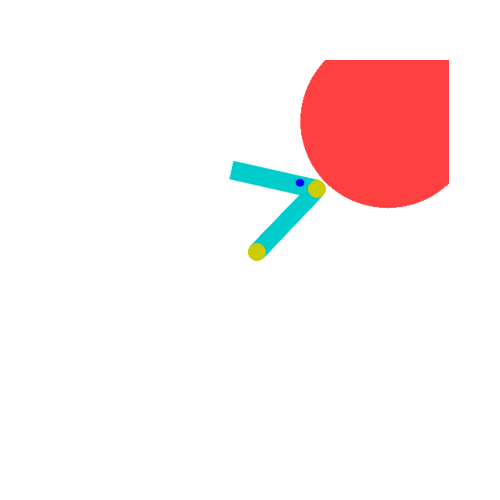}
         \caption{}
     \end{subfigure}
               \hfill
     \begin{subfigure}{0.24\textwidth}
         \centering
         \includegraphics[width=\textwidth]{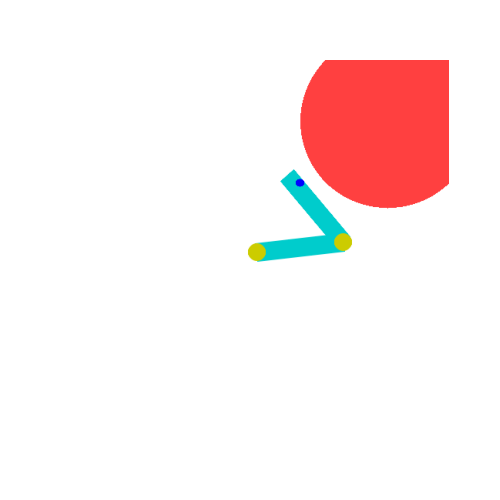}
         \caption{}
     \end{subfigure}
        \caption{Trajectory generated by the diffusion planner with value-function guide.}
        \label{fig:traj_value_only_extra3}
\end{figure*}

\begin{figure*}[h!]
     \centering
     \begin{subfigure}{0.24\textwidth}
         \centering
         \includegraphics[width=\textwidth]{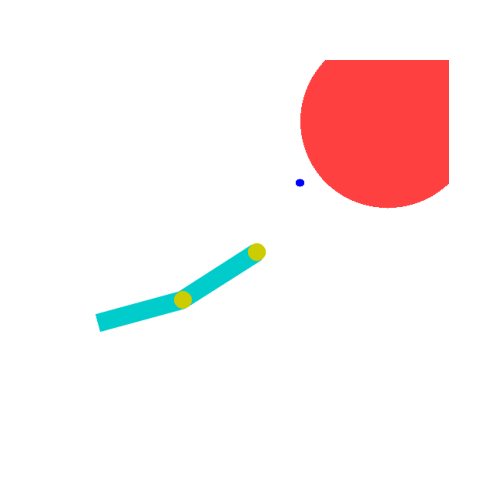}
         \caption{}
     \end{subfigure}
     \hfill
     \begin{subfigure}{0.24\textwidth}
         \centering
         \includegraphics[width=\textwidth]{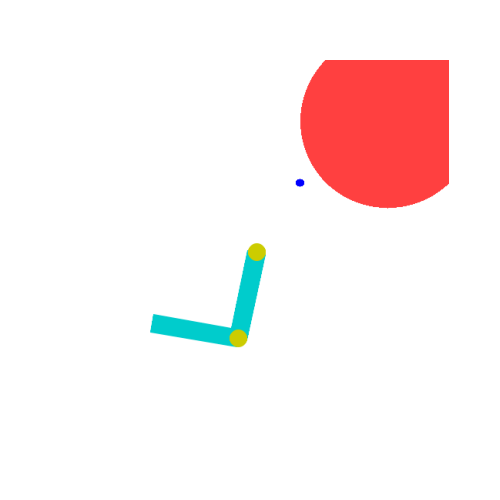}
         \caption{}
     \end{subfigure}
          \hfill
     \begin{subfigure}{0.24\textwidth}
         \centering
         \includegraphics[width=\textwidth]{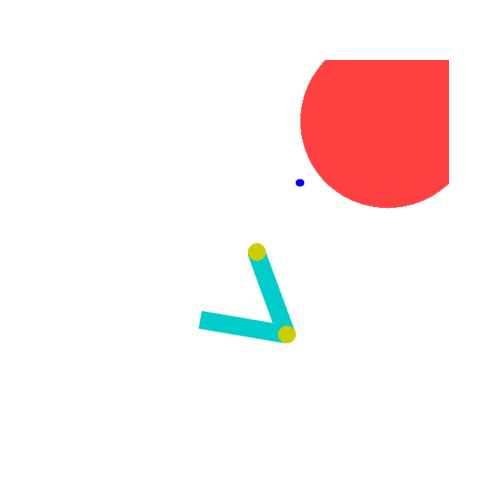}
         \caption{}
     \end{subfigure}
          \begin{subfigure}{0.24\textwidth}
         \centering
         \includegraphics[width=\textwidth]{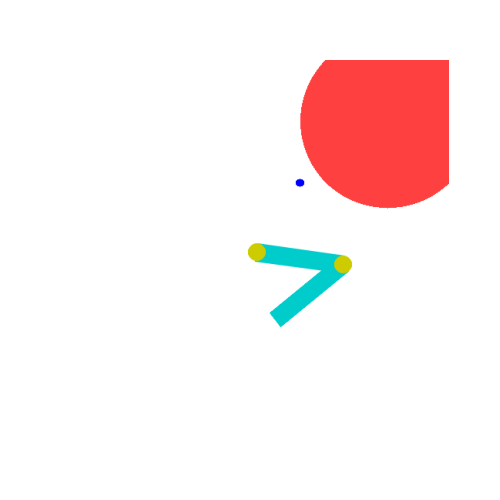}
         \caption{}
     \end{subfigure}
     \hfill
     \begin{subfigure}{0.24\textwidth}
         \centering
         \includegraphics[width=\textwidth]{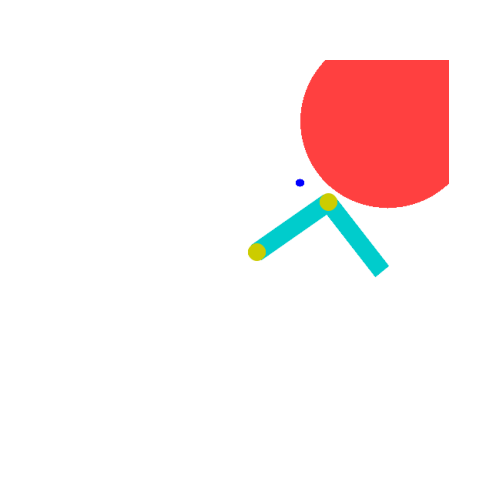}
         \caption{}
     \end{subfigure}
          \hfill
     \begin{subfigure}{0.24\textwidth}
         \centering
         \includegraphics[width=\textwidth]{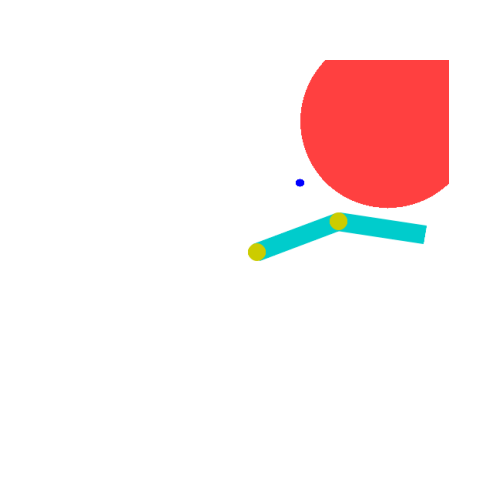}
         \caption{}
     \end{subfigure}
          \begin{subfigure}{0.24\textwidth}
         \centering
         \includegraphics[width=\textwidth]{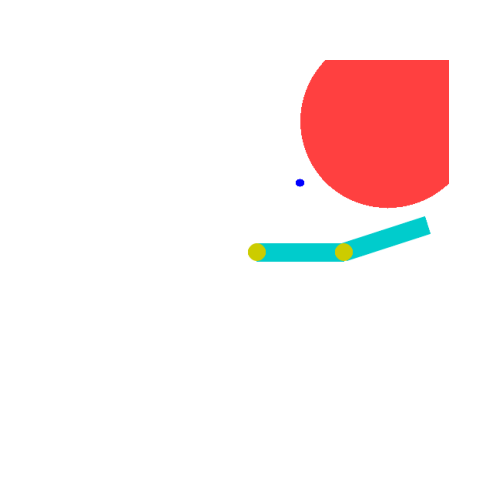}
         \caption{}
     \end{subfigure}
     \hfill
     \begin{subfigure}{0.24\textwidth}
         \centering
         \includegraphics[width=\textwidth]{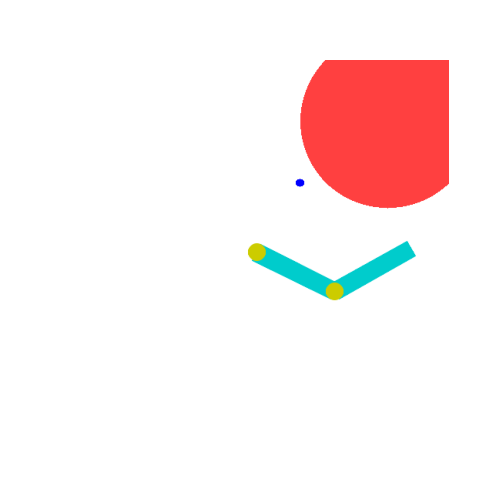}
         \caption{}
     \end{subfigure}
     \hfill
     \begin{subfigure}{0.24\textwidth}
         \centering
         \includegraphics[width=\textwidth]{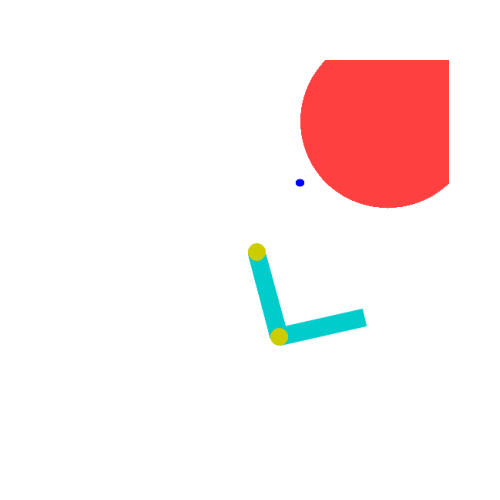}
         \caption{}
     \end{subfigure}
          \hfill
     \begin{subfigure}{0.24\textwidth}
         \centering
         \includegraphics[width=\textwidth]{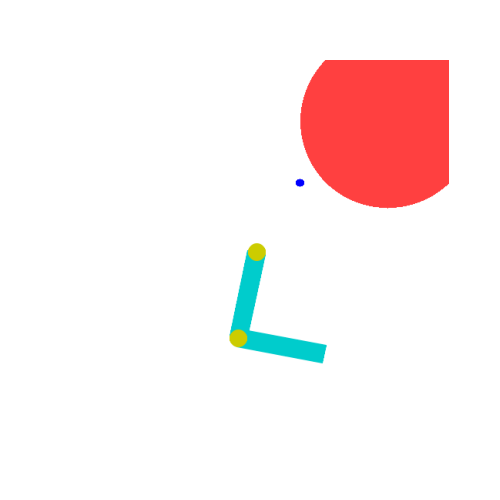}
         \caption{}
     \end{subfigure}
          \begin{subfigure}{0.24\textwidth}
         \centering
         \includegraphics[width=\textwidth]{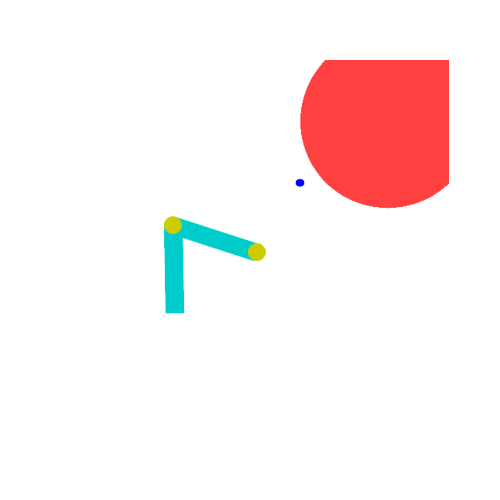}
         \caption{}
     \end{subfigure}
     \hfill
     \begin{subfigure}{0.24\textwidth}
         \centering
         \includegraphics[width=\textwidth]{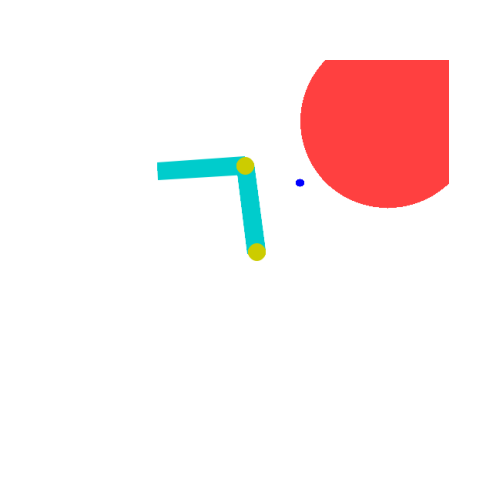}
         \caption{}
     \end{subfigure}
     \hfill
     \begin{subfigure}{0.24\textwidth}
         \centering
         \includegraphics[width=\textwidth]{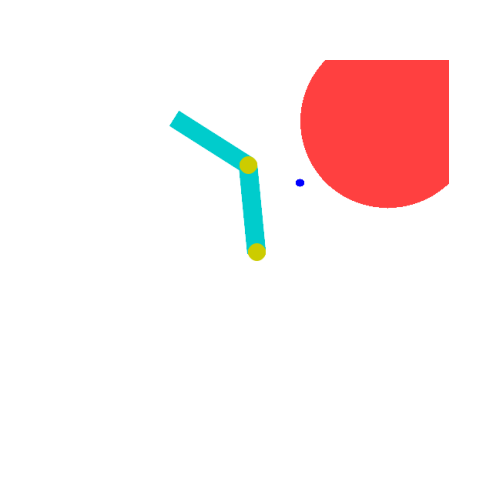}
         \caption{}
     \end{subfigure}
          \hfill
     \begin{subfigure}{0.24\textwidth}
         \centering
         \includegraphics[width=\textwidth]{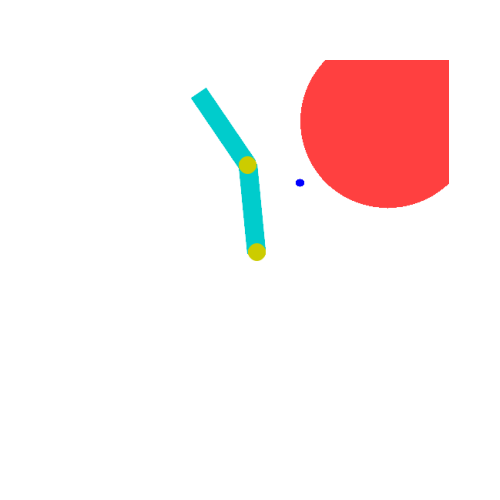}
         \caption{}
     \end{subfigure}
          \begin{subfigure}{0.24\textwidth}
         \centering
         \includegraphics[width=\textwidth]{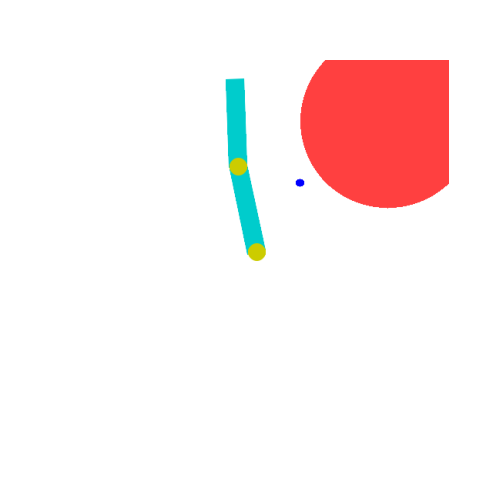}
         \caption{}
     \end{subfigure}
     \hfill
     \begin{subfigure}{0.24\textwidth}
         \centering
         \includegraphics[width=\textwidth]{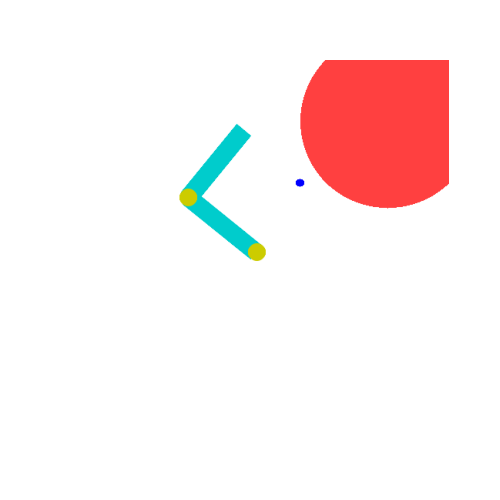}
         \caption{}
     \end{subfigure}
     \hfill
     \begin{subfigure}{0.24\textwidth}
         \centering
         \includegraphics[width=\textwidth]{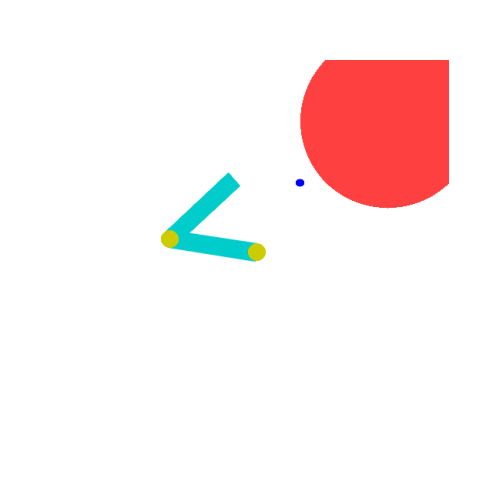}
         \caption{}
     \end{subfigure}
          \hfill
     \begin{subfigure}{0.24\textwidth}
         \centering
         \includegraphics[width=\textwidth]{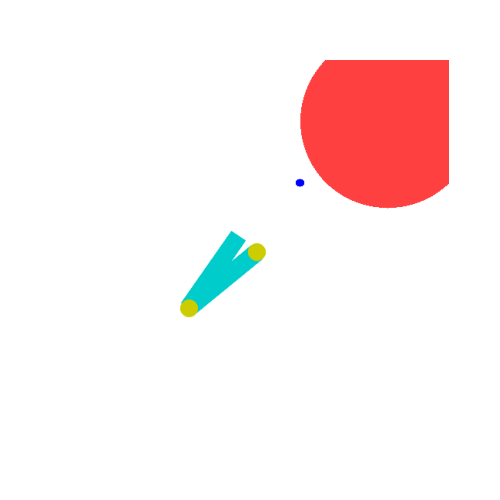}
         \caption{}
     \end{subfigure}
          \begin{subfigure}{0.24\textwidth}
         \centering
         \includegraphics[width=\textwidth]{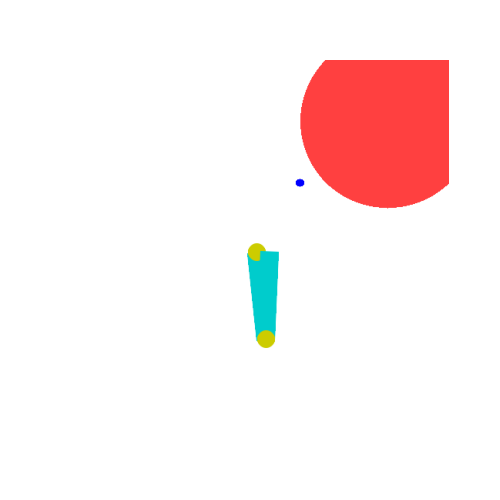}
         \caption{}
     \end{subfigure}
     \hfill
     \begin{subfigure}{0.24\textwidth}
         \centering
         \includegraphics[width=\textwidth]{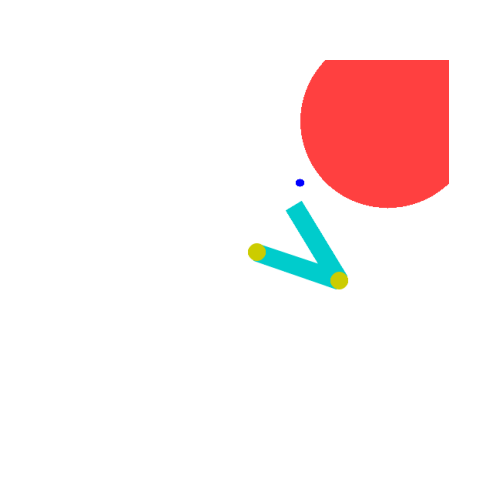}
         \caption{}
     \end{subfigure}
        \caption{Trajectory generated by the diffusion planner with value-function and safety-classifier guides.}
        \label{fig:traj_value_cbf_extra3}
    \end{figure*}


\begin{figure*}[h!]
     \centering
     \begin{subfigure}{0.24\textwidth}
         \centering
         \includegraphics[width=\textwidth]{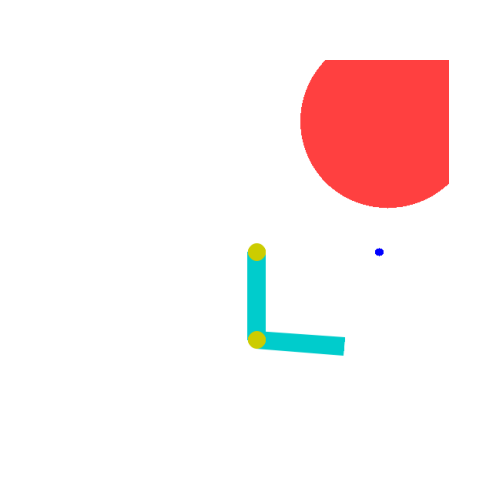}
         \caption{}
     \end{subfigure}
     \hfill
               \begin{subfigure}{0.24\textwidth}
         \centering
         \includegraphics[width=\textwidth]{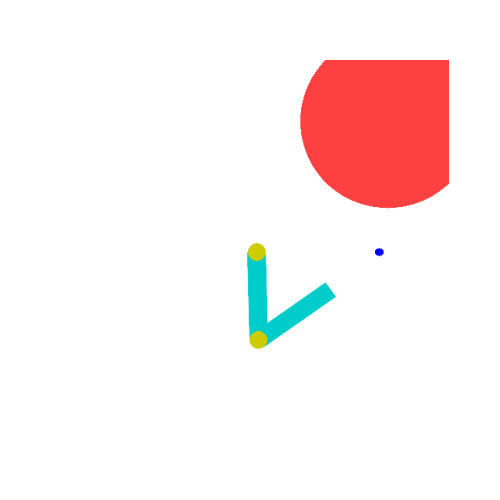}
         \caption{}
     \end{subfigure}
     \hfill
     \begin{subfigure}{0.24\textwidth}
         \centering
         \includegraphics[width=\textwidth]{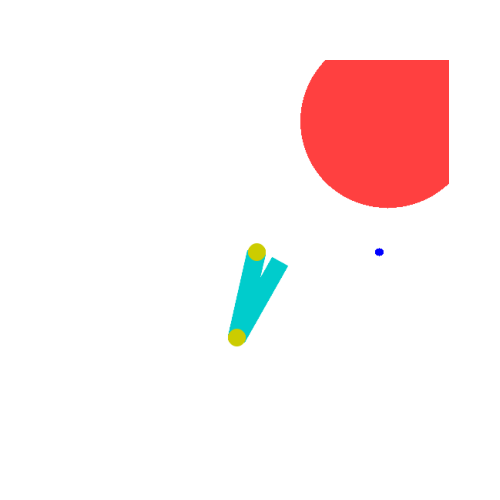}
         \caption{}
     \end{subfigure}
     \hfill
     \begin{subfigure}{0.24\textwidth}
         \centering
         \includegraphics[width=\textwidth]{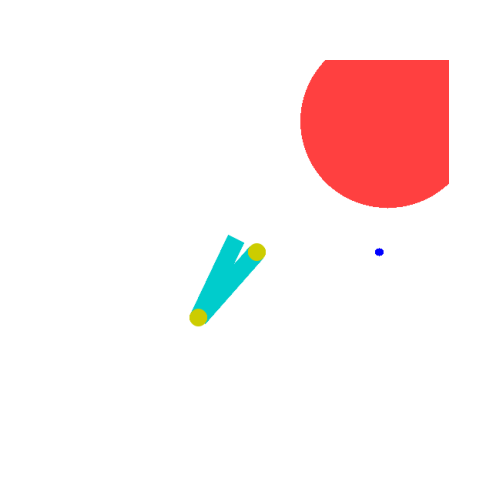}
         \caption{}
     \end{subfigure}
          \begin{subfigure}{0.24\textwidth}
         \centering
         \includegraphics[width=\textwidth]{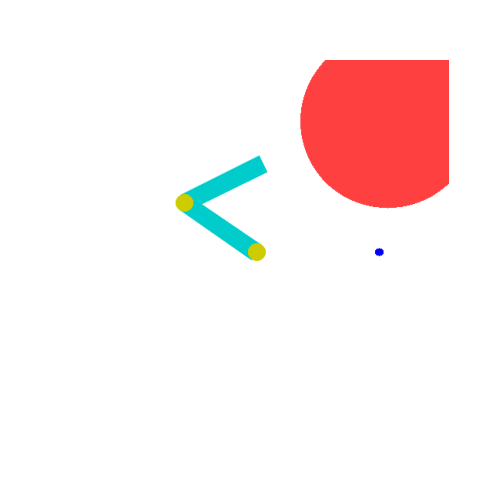}
         \caption{}
     \end{subfigure}
     \hfill
               \begin{subfigure}{0.24\textwidth}
         \centering
         \includegraphics[width=\textwidth]{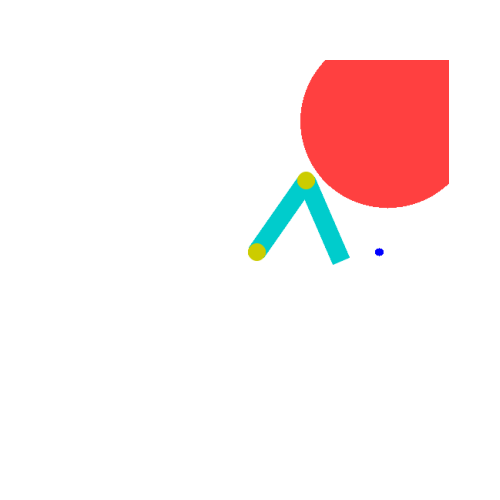}
         \caption{}
     \end{subfigure}
     \hfill
     \begin{subfigure}{0.24\textwidth}
         \centering
         \includegraphics[width=\textwidth]{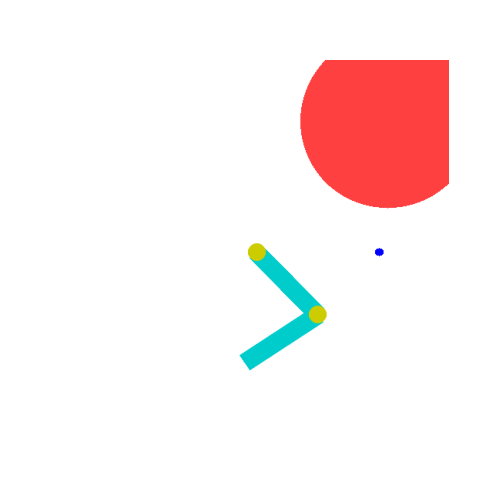}
         \caption{}
     \end{subfigure}
     \hfill
     \begin{subfigure}{0.24\textwidth}
         \centering
         \includegraphics[width=\textwidth]{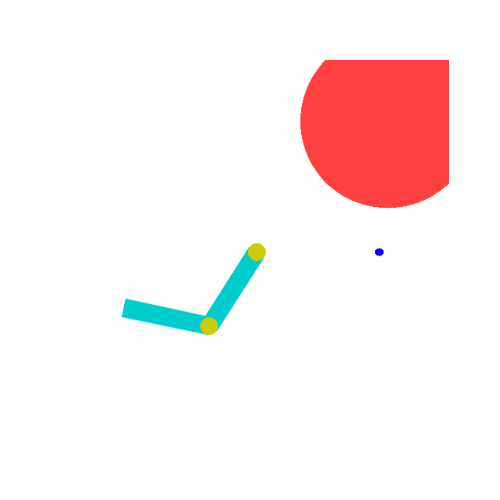}
         \caption{}
     \end{subfigure}
          \begin{subfigure}{0.24\textwidth}
         \centering
         \includegraphics[width=\textwidth]{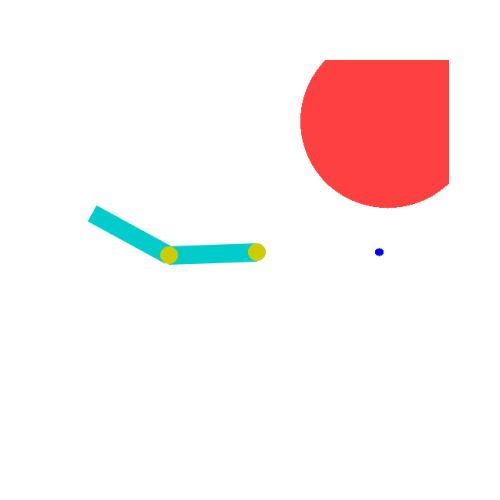}
         \caption{}
     \end{subfigure}
     \hfill
               \begin{subfigure}{0.24\textwidth}
         \centering
         \includegraphics[width=\textwidth]{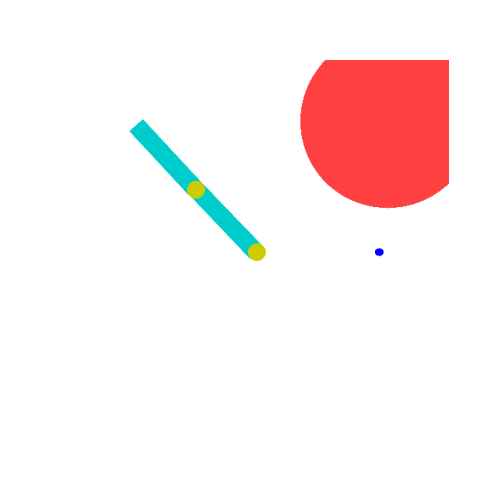}
         \caption{}
     \end{subfigure}
     \hfill
     \begin{subfigure}{0.24\textwidth}
         \centering
         \includegraphics[width=\textwidth]{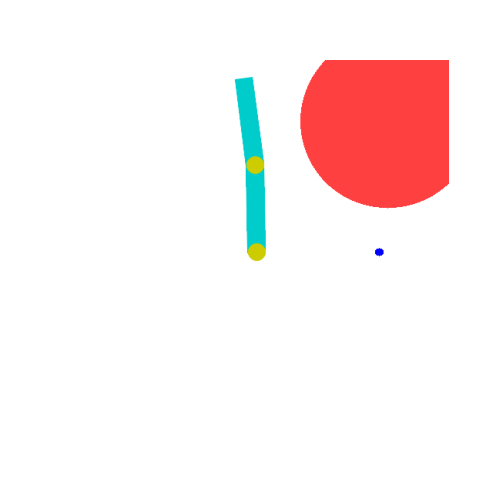}
         \caption{}
     \end{subfigure}
     \hfill
     \begin{subfigure}{0.24\textwidth}
         \centering
         \includegraphics[width=\textwidth]{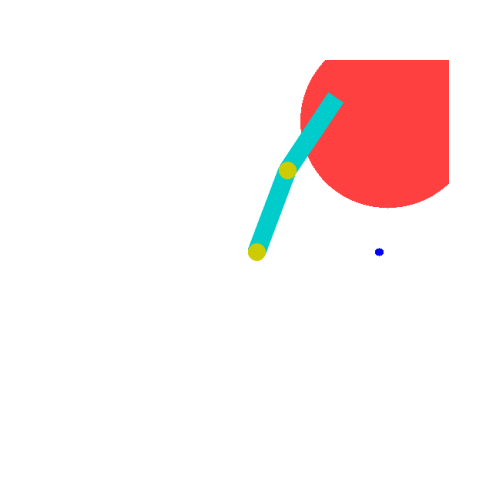}
         \caption{}
     \end{subfigure}
          \begin{subfigure}{0.24\textwidth}
         \centering
         \includegraphics[width=\textwidth]{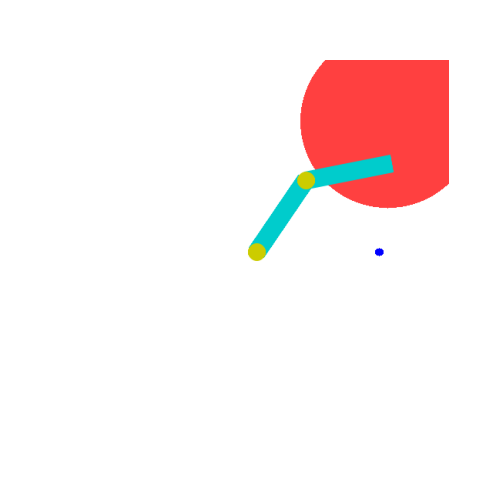}
         \caption{}
     \end{subfigure}
               \begin{subfigure}{0.24\textwidth}
         \centering
         \includegraphics[width=\textwidth]{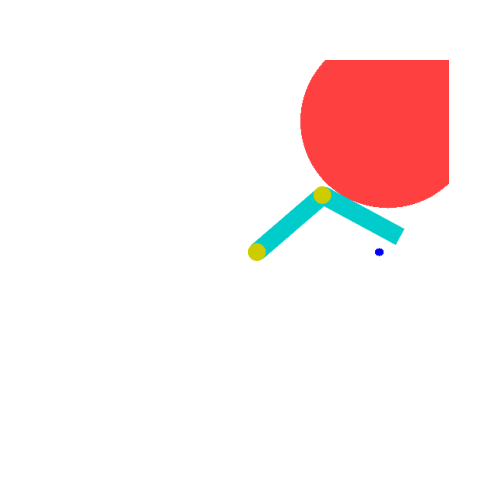}
         \caption{}
     \end{subfigure}
        \caption{Trajectory generated by the diffusion planner with value-function guide.}
        \label{fig:traj_value_only_extra4}
\end{figure*}

\begin{figure*}[h!]
     \centering
     \begin{subfigure}{0.24\textwidth}
         \centering
         \includegraphics[width=\textwidth]{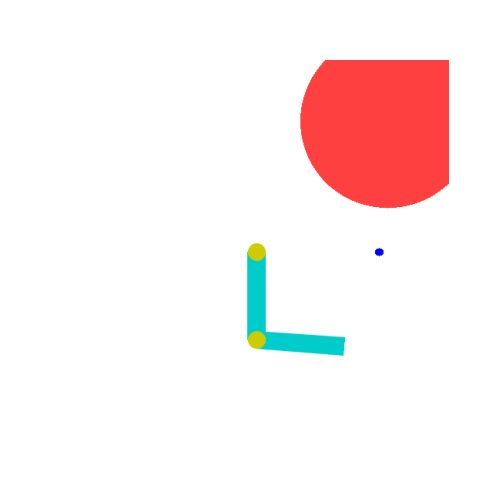}
         \caption{}
     \end{subfigure}
     \hfill
     \begin{subfigure}{0.24\textwidth}
         \centering
         \includegraphics[width=\textwidth]{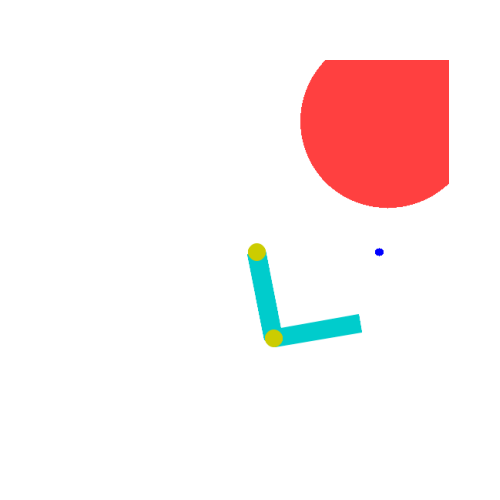}
         \caption{}
     \end{subfigure}
     \hfill
     \begin{subfigure}{0.24\textwidth}
         \centering
         \includegraphics[width=\textwidth]{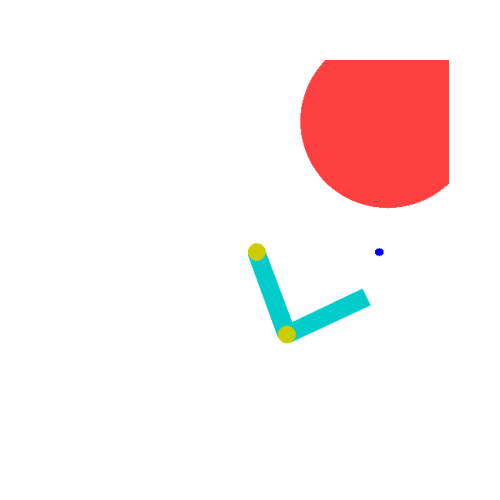}
         \caption{}
     \end{subfigure}
          \hfill
     \begin{subfigure}{0.24\textwidth}
         \centering
         \includegraphics[width=\textwidth]{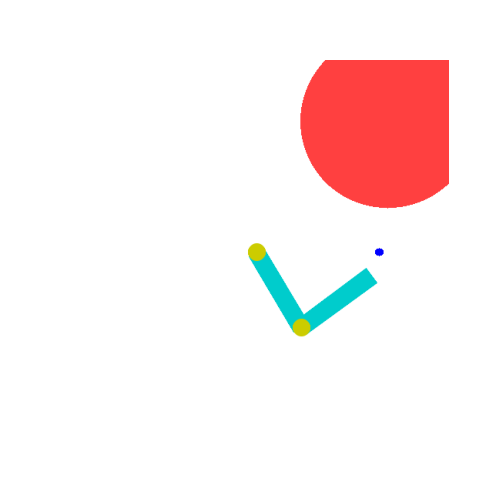}
         \caption{}
     \end{subfigure}
        \caption{Trajectory generated by the diffusion planner with value-function and safety-classifier guides.}
        \label{fig:traj_value_cbf_extra4}
    \end{figure*}

    
\end{document}